\documentclass[journal]{IEEEtran}

\ifCLASSINFOpdf
\else
\fi

  \usepackage[colorlinks,hyperindex,breaklinks]{hyperref}

\usepackage{epsfig}
\usepackage{graphicx}
\usepackage{amsthm,amsmath}
\usepackage{amsfonts}

\usepackage{amssymb}
\usepackage{mathrsfs}
\usepackage{subfigure}

\usepackage{color}
\usepackage{threeparttable}
\usepackage{bm}

\usepackage{algorithm}
\usepackage{algorithmic}
\usepackage{amsfonts,amssymb}
\usepackage{dsfont}

\newcommand{\myparagraph}[1]{\vspace{0.1em}\noindent\textbf{#1}}
\newcommand{\ie}{\textit{i}.\textit{e}.}
\newcommand{\eg}{\textit{e}.\textit{g}.}

\begin{document}
%
\title{Anchor Retouching via Model Interaction for Robust Object Detection in Aerial Images}
%
%
%

\author{Dong Liang$^{\dag}$,~\IEEEmembership{Member,~IEEE,}
        Qixiang~Geng,
        Zongqi~Wei,
        Dmitry A. Vorontsov,
        Ekaterina L. Kim, \\
       Mingqiang Wei,~\IEEEmembership{Senior Member,~IEEE},
          and Huiyu Zhou
        
\thanks{D. Liang, Q. Geng, Z. Wei, and Mingqiang Wei are with the College of Computer Science and Technology, Nanjing University of Aeronautics and Astronautics, MIIT Key Laboratory of Pattern Analysis and Machine Intelligence, Collaborative Innovation Center of Novel Software Technology and Industrialization, Nanjing 211106, China. E-mail: \{liangdong, gengqx, weizongqi, mingqiangw\}@nuaa.edu.cn}
\thanks{D. A. Vorontsov and E. L. Kim are with the National Research Lobachevsky State University of Nizhny Novgorod, Nizhny Novgorod, Russia.  E-mail: vorontsovda@mail.ru; kim@phys.unn.ru}
\thanks{H. Zhou is with the School of Computing and Mathematical Sciences, University of Leicester, Leicester LE1 7RH, United Kingdom. E-mail: hz143@leicester.ac.uk}
\thanks{$\dag$ Corresponding author: Dong Liang.}}

%
%


\maketitle

\begin{abstract}
Object detection has made tremendous strides in computer vision. Small object detection with appearance degradation is a prominent challenge, especially for aerial observations. To collect sufficient positive/negative samples for heuristic training, most object detectors preset region anchors in order to calculate Intersection-over-Union~(IoU) against the ground-truthed data. In this case, small objects are frequently abandoned or mislabeled. In this paper, we present an effective Dynamic Enhancement Anchor~(DEA) network to construct a novel training sample generator. Different from the other state-of-the-art techniques, the proposed network leverages a sample discriminator to realize interactive sample screening between an anchor-based unit and an anchor-free unit to generate eligible samples. Besides, multi-task joint training with a conservative anchor-based inference scheme enhances the performance of the proposed model while reducing computational complexity. The proposed scheme supports both oriented and horizontal object detection tasks. Extensive experiments on two challenging aerial benchmarks (\ie, DOTA and HRSC2016) indicate that our method achieves state-of-the-art performance in accuracy with moderate inference speed and computational overhead for training. On DOTA, our DEA-Net which integrated with the baseline of RoI-Transformer surpasses the advanced method by 0.40\% mean-Average-Precision~(mAP) for oriented object detection with a weaker backbone network (ResNet-101 \emph{vs} ResNet-152) and 3.08\% mean-Average-Precision~(mAP) for horizontal object detection with the same backbone. Besides, our DEA-Net which integrated with the baseline of ReDet achieves the state-of-the-art performance by 80.37\%. On HRSC2016, it surpasses the previous best model by 1.1\% using only 3 horizontal anchors. The source code and the training set are made publicly available at: \href{https://github.com/QxGeng/DEA-Net}{https://github.com/QxGeng/DEA-Net}
\end{abstract}


\section{Introduction}

\IEEEPARstart{O}{bject} detection is one of the fundamental and challenging problems in computer vision. Tremendous successes have been achieved on object detection with the development of deep Convolution Neural Networks~(DCNNs) in recent years. Different from the objects in natural scenes which are often captured from horizontal perspectives, aerial images are typically taken from a bird's-eye view at a high altitude, suggesting that objects in aerial images usually are of a small size and diverse orientations with complex background~\cite{xia2018dota}. A large number of detectors ~\cite{ding2019learning,yang2019scrdet,ming2020dynamic,yang2020dense} have been designed for aerial observations, most of which are based on a two-stage
detector~(\eg, Fast R-CNN~\cite{girshick2015fast} and Faster R-CNN~\cite{ren2016faster})
or a one-stage detector~(\eg, RetinaNet~\cite{lin2017focal} and YOLO~\cite{redmon2016you}).

\begin{figure}[t]
	\centering
	\subfigure[]{\includegraphics[width=.48\linewidth]{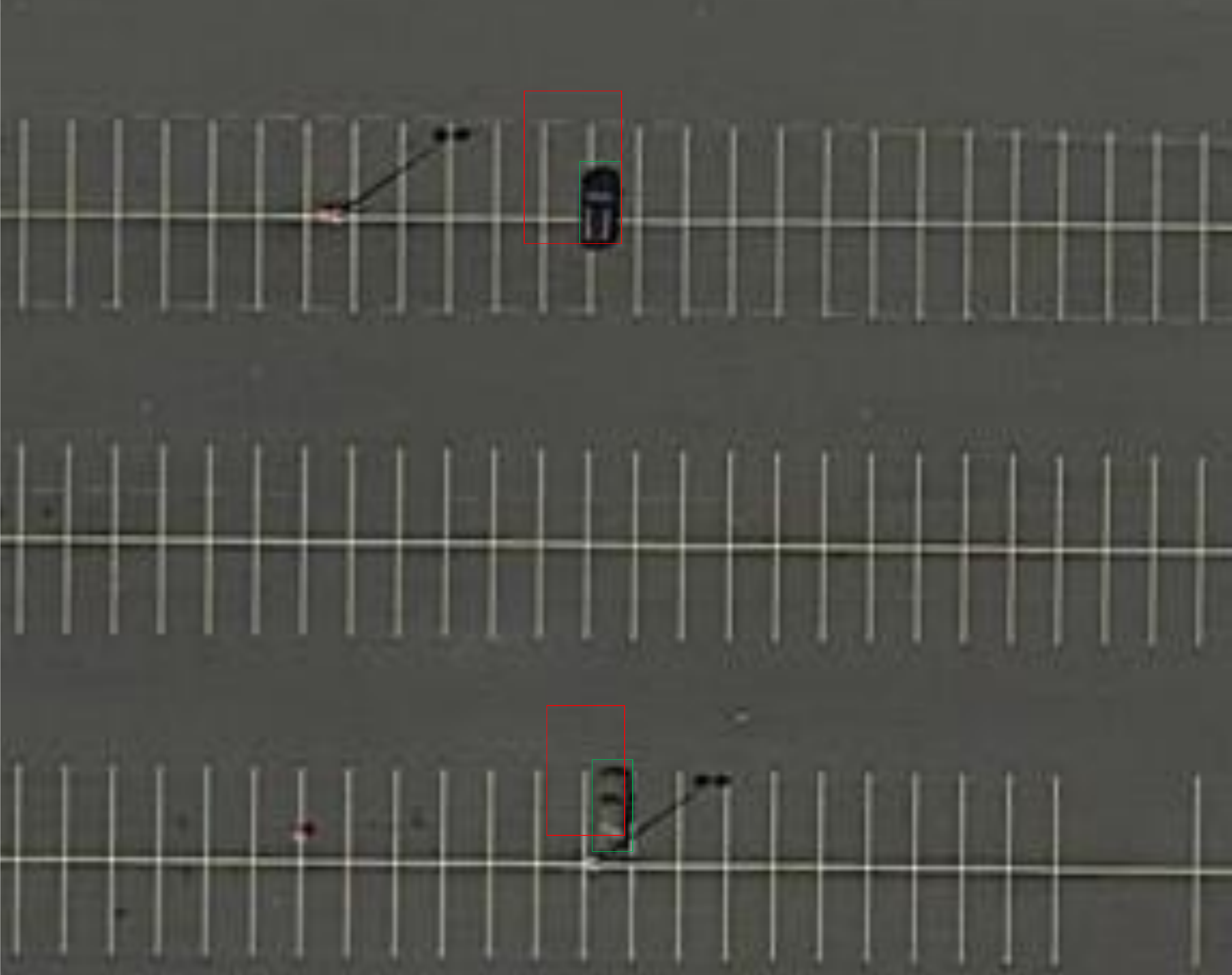}}
	\subfigure[]{\includegraphics[width=.48\linewidth]{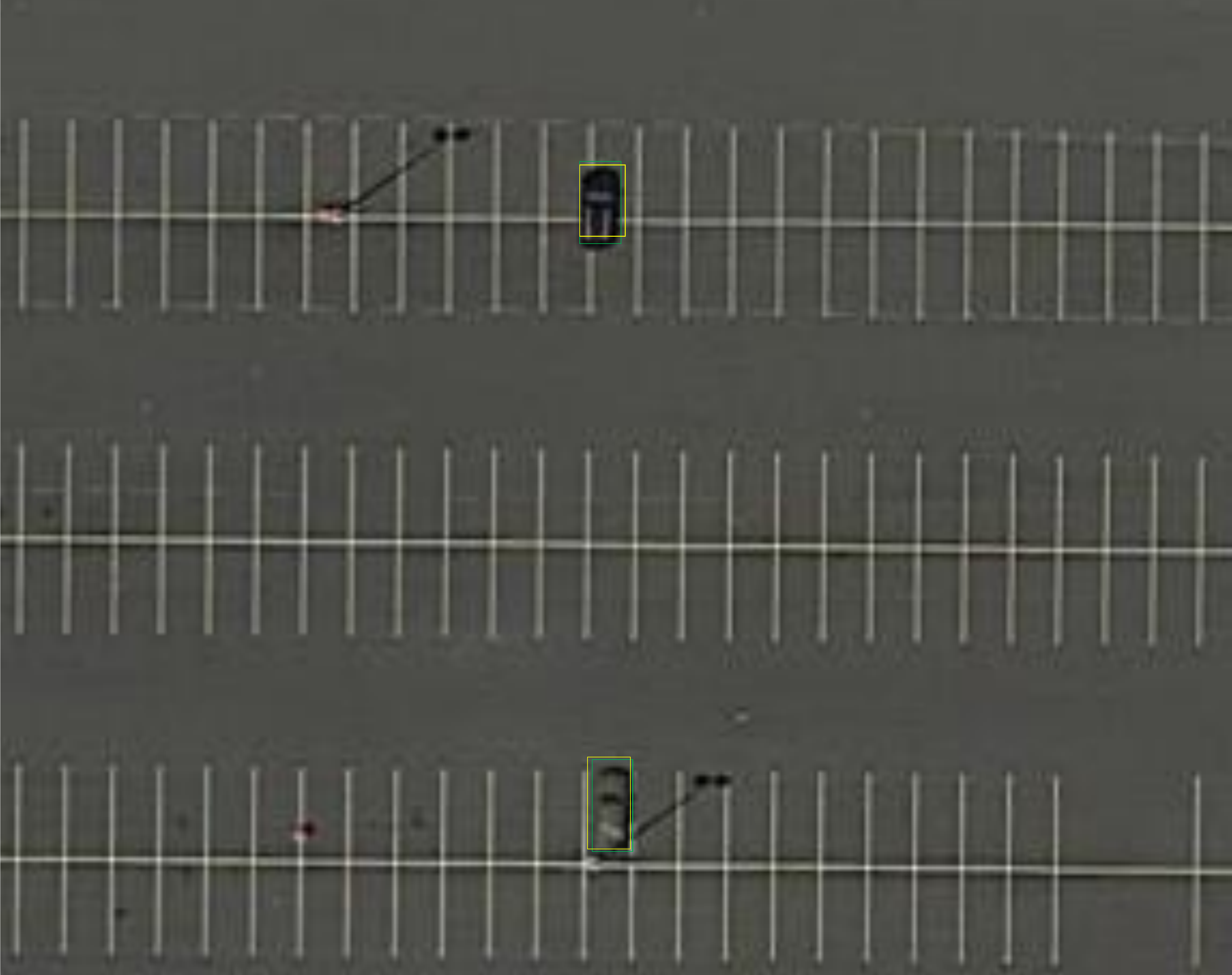}}
	\subfigure[]{\includegraphics[width=.48\linewidth]{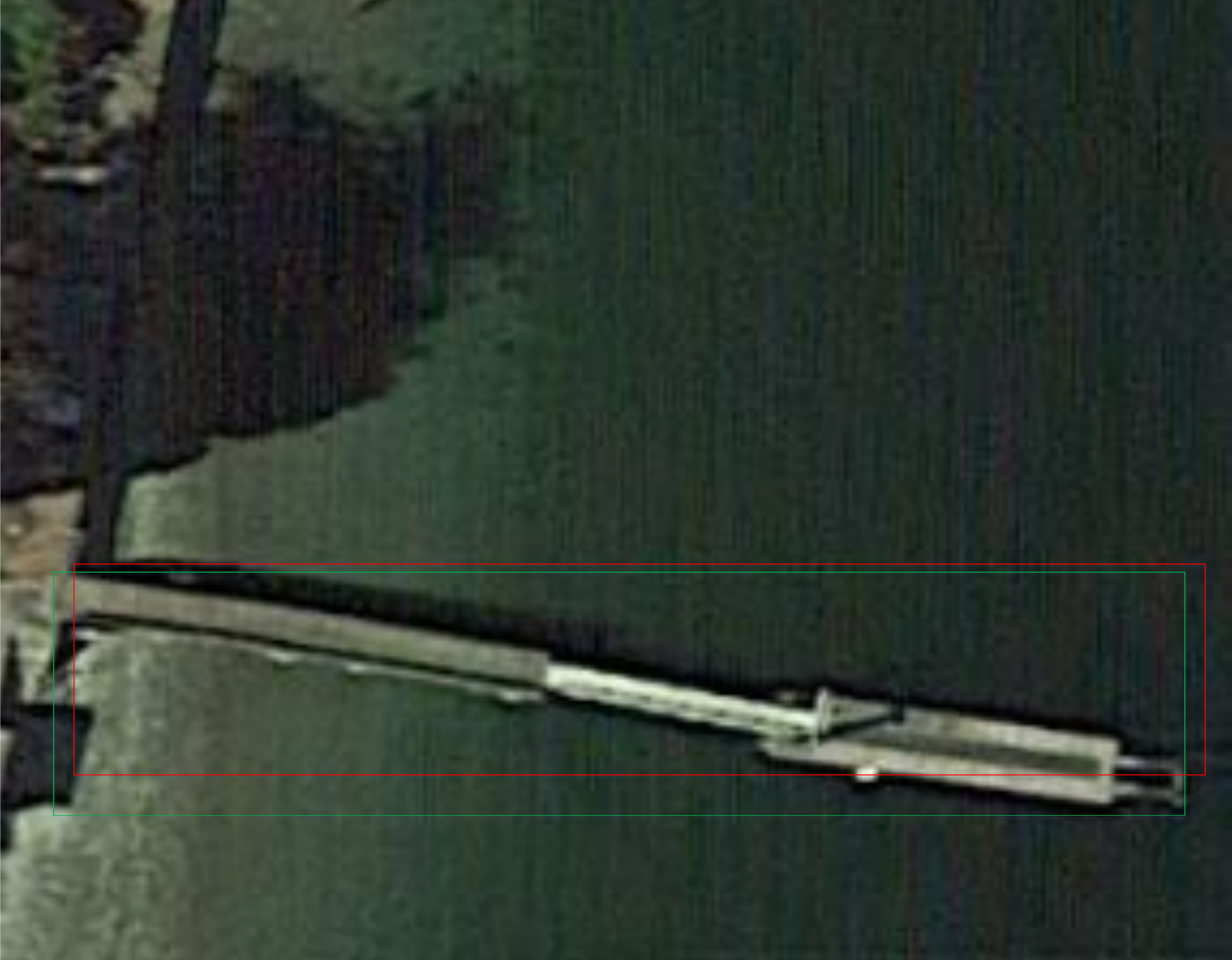}}
	\subfigure[]{\includegraphics[width=.48\linewidth]{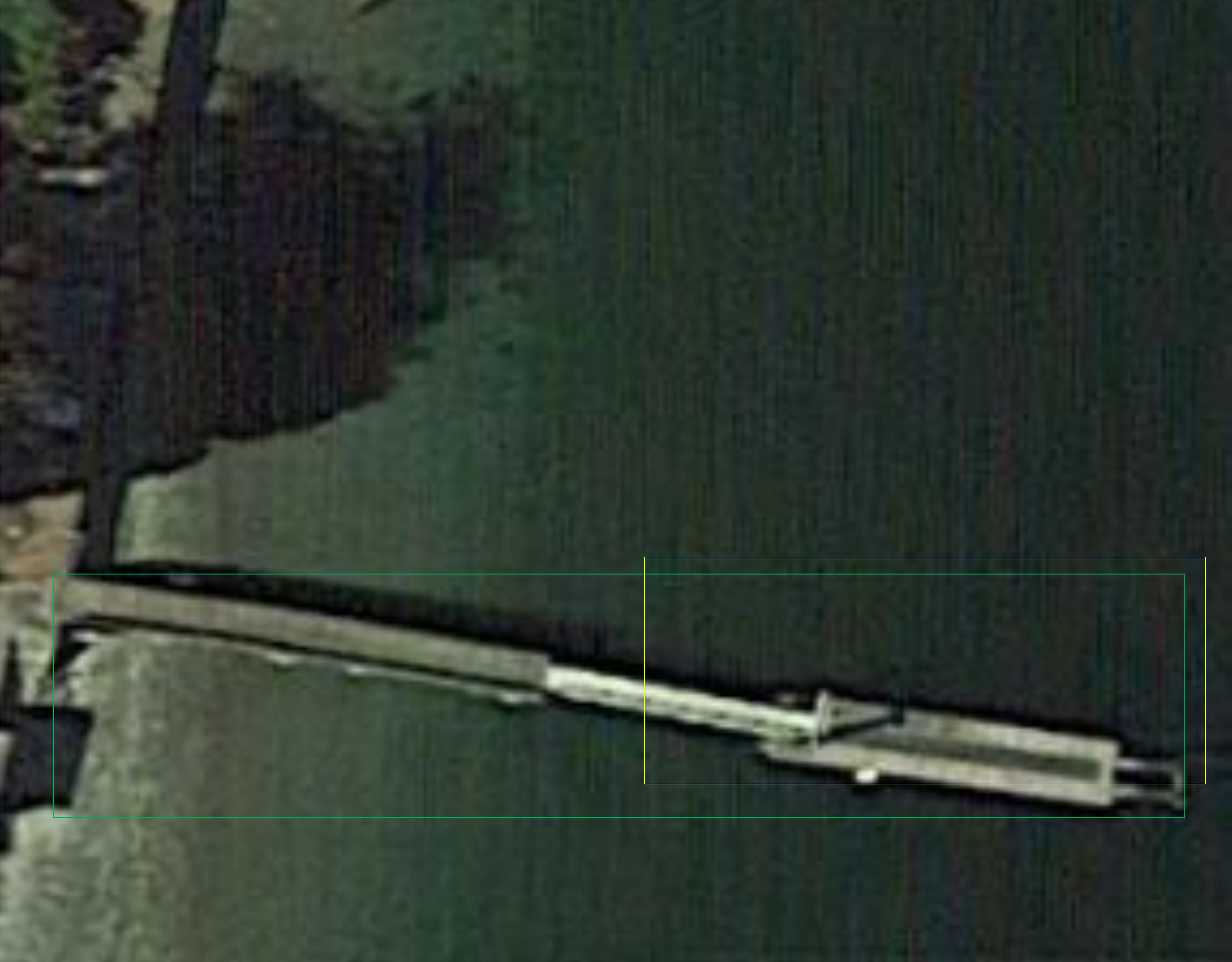}}
	\caption{Comparison of anchor proposals~(a) and (c) in an anchor-based  method~(e.g. Faster-RCNN~\cite{ren2016faster}) and regression bounding-boxes~(b) and (d) in an anchor-free method~(e.g. FCOS~\cite{tian2019fcos}). Anchor-free regression bounding-boxes ({\color{yellow} yellow} boxes) have better consistency in small object detection than the anchor proposals ({\color{red} red} boxes), but for large objects with large aspect ratios, the anchor proposals have better consistency. {\color{green} Green} boxes are the ground truth.}
	\label{compare}
\end{figure}
Region anchors are designed as the regression references and the classification candidates to predict the proposals in two-stage detectors or final bounding boxes in one-stage detectors. Most of the anchor-based detectors utilize a uniform anchoring scheme, and then positive and negative samples are selected through Intersection-over-Union~(IoU) with ground-truth. For example, the anchor boxes with $ IoU > 0.5 $ are treated as positive samples and $IoU < 0.3$ as negative samples~\cite{ren2016faster}. In practice, such a strategy may cause two main problems:  

\myparagraph{(i) Anchor quantization errors and noisy training samples} 

We take Faster-RCNN \cite{ren2016faster} as an example. If the base anchor size is set to 32 and the IoU threshold is set to 0.5, objects with area $< 32^2\times 0.5$ (512 pixels) will be excluded from the positive proposals. As shown in Figure \ref{compare}~(a), the anchor box (red) and the ground truth (green) have a large quantization discrepancy. This discrepancy will lead to much confusion for both box localization and classification. On the other hand, empirical evidence shows that objects with an area $< 512$ pixels occupy approximately 30$\%$ of an aerial image of DOTA~\cite{xia2018dota}, where inaccurate anchor-boxes or misclassification of the samples lead to unstable convergence of the model. 

\myparagraph{(ii) Mismatch between the pyramid levels and the samples} 

This is based on the consensus that upper feature maps have more semantic information suitable for detecting big instances whereas lower feature maps have more fine-grained details suitable for detecting small instances. Integrated within a feature pyramid, large anchor proposals are typically associated with upper feature maps, and small anchor proposals are associated with lower feature maps. Inaccurate bounding boxes with large background areas would cause a mismatch between the feature pyramid levels and the training samples, largely affecting the model training. In other words, the selected feature level to train each sample may not be correct.

To deal with these issues, image feature pyramids with more levels can be used to better detect small objects. Another common solution is to enlarge the quantity of the anchors with diverse sizes and aspect ratios. These two solutions have evident drawbacks -- both of them lead to significant computational overhead, especially when processing large-scale aerial images or training the network with a heavy backbone.  

As shown in Figure \ref{compare}~(b), the regression bounding-boxes in an anchor-free detector (\eg ~FCOS~\cite{tian2019fcos}) can be potentially leveraged as positive region proposals because they are free from anchor quantization errors. On the other hand, compared with the anchor-based detectors, the anchor-free detectors usually fail to generate an accurate bounding box when the objects are of a large size and an extreme aspect ratio~\cite{law2018cornernet,zhou2019objects,tian2019fcos}, just like the example shown in Figure \ref{compare}~(d). Most anchor-based detector (including the baseline Faster R-CNN) regresses from the anchor box with four offsets between the anchor box and the object box, while FCOS regresses from one point with four distances to the bound of the object. It means that for a positive sample, the regression's starting status of Faster R-CNN is a box while FCOS is a point. The box itself contains prior of the shape, and the regression is only the two small offset of $X$ and $Y$ directions. In contrast, FCOS needs to independently return the offsets of +/-$X$ and $Y$, four directions, from a start point, without any shape prior. The regression error in any direction would greatly affect the shape of the box. Especially for large objects with a large aspect ratio. This observation has been reported in \cite{2020Corner}.

For the anchor-based approaches, the anchor quantization errors can be ignored for large objects. The anchor boxes are designed to discretize all possible instance boxes into a finite number of boxes with predefined locations, scales, and aspect ratios. We need more anchors with a smaller size and denser layouts or more angles in arbitrary-oriented detection to cover small objects, which may lead to extensive computation cost and imbalanced problems of positive and negative samples. Achieving spatial alignment with small ground-truth objects is challenging and prone to the miss of the corresponding positive anchors based on this strategy. 

\begin{figure}
	\centering
	\includegraphics[width=1\linewidth]{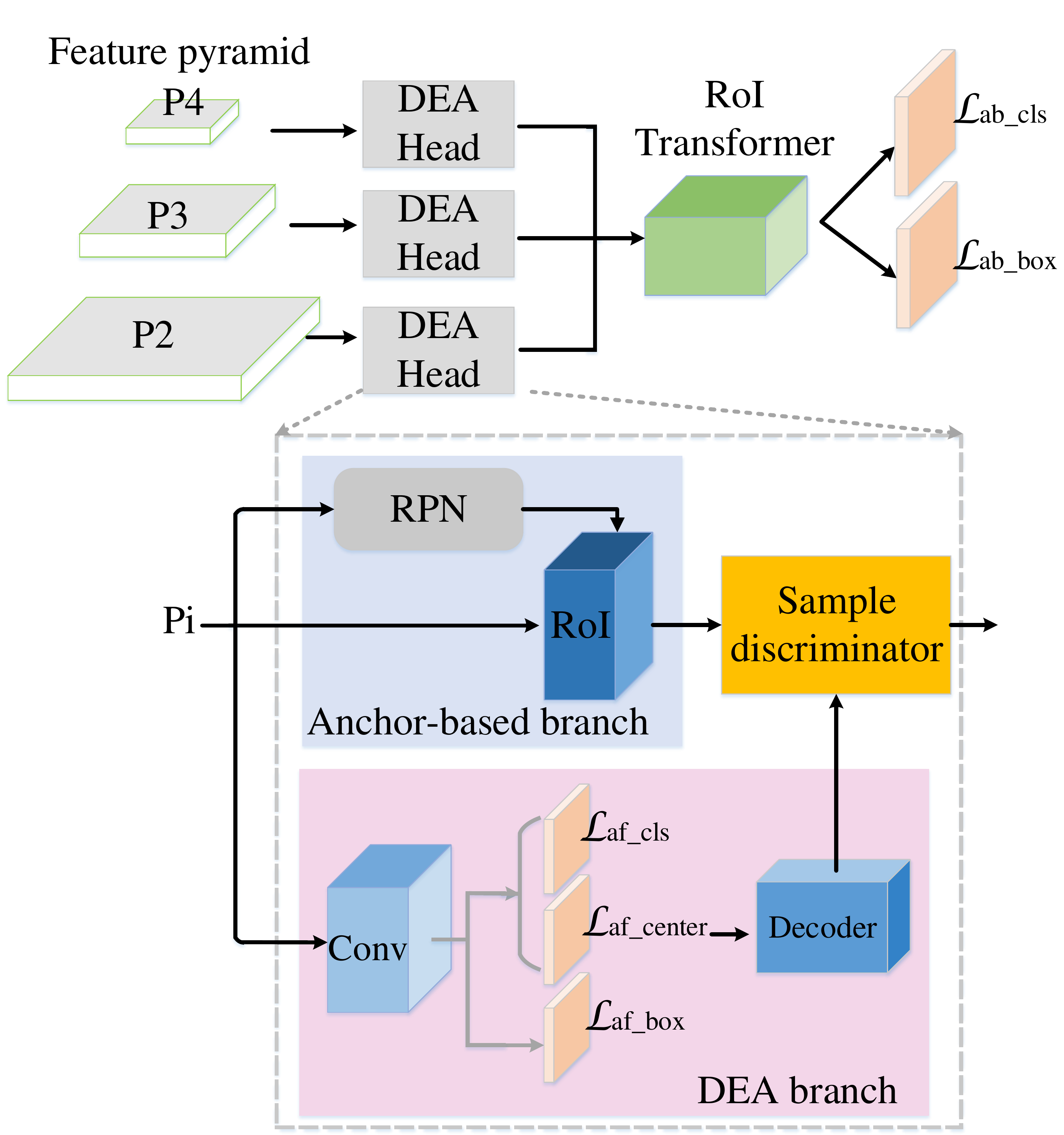}
	\caption{The architecture of our proposed Dynamic Enhancement Anchor Network~(DEA-Net) for oriented object detection. The DEA head serves each level of the feature pyramid to generate higher-quality training samples, including an anchor-based module, an anchor-free module, and a sample discriminator. RoI transformer \cite{ding2019learning} processes the horizontal and the oriented proposals.}
	\label{structure}
\end{figure}

Inspired by the above observations, in this paper, we propose an effective Dynamic Enhancement Anchor (DEA) network to enhance the learning of small objects efficiently. The overall architecture is shown in Figure~\ref{structure}. The DEA head serves each level of the feature pyramid consisting of an anchor-based module, an anchor-free module, and a sample discriminator. The sample discriminator merges the complementary anchor-based and the anchor-free proposals and generates representative and informative samples with accurate locations and sizes, while avoiding positive/negative confusion. Besides, multi-task joint training with a conservative anchor-based inference scheme enhances the performance of the model while avoiding complexity augmentation. We conduct extensive experiments on both oriented and horizontal object detection tasks. Experiments on the aerial image benchmarks DOTA~\cite{xia2018dota} and HRSC2016~\cite{lb2017high} show that our proposed DEA-Net makes substantial improvement, compared to the baseline methods, and achieves state-of-the-art performance in accuracy (\ie, $80.37\%$ mean-Average-Precision~(mAP) (+0.14\%) and $90.56\%$ mean-Average-Precision~(mAP) (+1.10\%)) for oriented object detection tasks. Besides, experiments on DOTA~\cite{xia2018dota} for horizontal object detection  achieves state-of-the-art performance in accuracy with $78.43\%$ mean-Average-Precision~(mAP) (+3.08\%). By combining the anchor-based and anchor-free branch efficiently, our method maintains a fair inference speed and  training computational overhead. 

To our knowledge, this is the first time to simultaneously consider the impact of both the object’s scale and aspect ratio, and then distinguish and process them separately for training.
In summary, our main contributions consist of: (1) An  effective sample generator based on DEA head to enhance the performance of detecting small objects by combining the advantages of anchor-base and anchor-free models. (2) A novel and robust DEA-Net, which can achieve the start-of-the-art oriented and horizontal object detection performance in aerial images. 

The remainder of this paper is organized as follows. In Section II, we discuss the related work. In Section III, we describe the proposed method in detail. The experimental results are presented and discussed in Section IV, and the conclusions and future work are given in Section V.

\section{Related Work}
\subsection{Anchor-based and anchor-free models} 
The current mainstream detectors can be divided into two categories: (1) Anchor-based methods \cite{girshick2015fast, ren2016faster,lin2017focal,redmon2016you} and (2) anchor-free methods ~\cite{law2018cornernet, zhou2019objects,tian2019fcos}. 
In anchor-based methods, the network is trained to regress the offsets between the anchors and ground truth bounding boxes. However, these methods take advantages of the task-oriented settings of anchors, leading to complex parameter tuning. Moreover, since the scales and aspect ratios of anchors are fixed, it has the difficulty to
handle the objects with large shape variations, especially for small objects.  Anchor-free methods directly regress the bounding box without using preset anchors. They usually have a streamlined network structure due to discarding of the dense anchors. However, they also meet difficulties in learning large variations of the bounding boxes without prior knowledge. DETR~\cite{DETR} utilizes self-attention to build a novel detection architecture, whose detection precision can compete against those of the two-stage object detectors, but it has the weakness in detecting small objects with high computational overheads in the published literature. Due to the dilemma of the above methods, an emerging line of work attempts to design a detector by combining anchor-based and anchor-free methods. GA-RPN~\cite{wang2019region} constructs a region proposal network in an anchor-free manner to predict the proposals for Faster R-CNN. FSAF~\cite{zhu2019feature} attaches an anchor-free module at each feature pyramid level to select appropriate features of each object for RetinaNet. SFace~\cite{wang2018sface} attaches an anchor-free module to an anchor-based detector and combines the outputs of two modules to improve the performance of the detector. Different from the other state of the arts such as~\cite{wang2019region}, \cite{zhu2019feature} and ~\cite{wang2018sface}, we focus on the collaboration of anchor-based and anchor-free methods from the perspective of sample discrimination. It makes interactive sample screening invulnerable to the diversity of the scale distributions

\subsection{Improved anchor-based detection models} 
The recent improvement of the anchor selection strategies mainly focuses on two aspects: (1) Weight the predicted anchor boxes to distinguish the potential importance and quality differences. MetaAnchor~\cite{MetaAnchor} and soft anchor-point object detection (SAPD)~\cite{Zhu2020SoftAO} belong to this type. MetaAnchor directly weights the generated anchors, while SAPD leverages both soft-weighted anchor points and soft-selected pyramid levels. (2) Propose a refining anchor box assignment strategy. For example, Dynamic Anchor Feature Selection (DAFS) ~\cite{9009487} uses an Anchor Refinement Module (ARM) to adjust the locations and sizes of anchors, and filter out negative anchors and then select new pixels in a feature map for each refined anchor. 
In ~\cite{ming2020dynamic}, the authors define the dynamic anchor with a matching degree to evaluate both spatial and feature alignment for anchor assignment. 
Nevertheless, all the above methods ignore the influence of the object’s scale and aspect ratio on anchor assignment. In remote sensing scenarios, for example, we observe that small objects are frequently abandoned or mislabeled due to the predefined anchor sampling interval and the Intersection-over-Union~(IoU) rule, which could potentially destroy the original sample distribution in the feature space. On the other hand, anchor-free based detectors often fail to generate an accurate bounding box for large objects with large aspect ratio. 
Different from ~\cite{MetaAnchor, Zhu2020SoftAO, 9009487, ming2020dynamic}, the significant differences of our method are (1) consider the impact of both the object’s scale and aspect ratio simultaneously, and (2) distinguish between different scales and aspect ratios, and then use appropriate strategies to process them separately. The anchor-free module is utilized to generate more positive samples of small objects which are ignored in the anchor-based module. For some objects of a large size and extreme aspect ratios, we preserve the anchors in the anchor-based module which have higher IoUs as the positive samples. 
 Compared with the existing methods (weighting the anchor ~\cite{MetaAnchor, Zhu2020SoftAO} or refinement assignment strategy ~\cite{9009487, ming2020dynamic}), our idea has also been experimentally proven to be effective. In particular, it is more suitable for object detection in remote sensing images, because in remote sensing images, small-sized (such as vehicles) and large-sized objects with extreme aspect ratios (such as ports, bridges) are very common. The work ATSS \cite{2020Bridging} is a comparative analysis of the Retinanet and FCOS, which proposes a general training strategy to serve them separately so this method naturally does not increase any overhead. In contrast, our method involves model design, a training scheme (relies on a sample discriminator for interactive sample screening and with multi-task joint training), and an inference scheme (a conservative anchor-based scheme to freezing the anchor-free branch to suppress computational complexity). Such a comprehensive scheme improves the performance in remote sensing scenarios. 
 
\subsection{Object detection in aerial images} Object detection in aerial images often faces a large number of small objects with arbitrary orientations in complex environments. Detecting objects with oriented bounding boxes is a non-trivial extension of horizontal object detection, which are mostly built on anchor-based detectors~\cite{ma2018arbitrary,liu2016ship,liao2018textboxes++, ding2019learning}. 
R$^3$Det \cite{yang2019r3det} adopts cascade regression to refine the predicted boxes.  DCL~\cite{yang2020dense} utilizes a densely coded label encoding mechanism for angle classification. SCRDet~\cite{yang2019scrdet} improves the performance of small objects by reducing the anchor strides to preset smaller and more anchors, which incurs extensive computational costs.  DAL~\cite{ming2020dynamic} utilizes a comprehensive scheme for spatial alignment, feature alignment ability, and regression uncertainty for label assignment. RoI Transformer ~\cite{ding2019learning} applies spatial transformations on RoIs
and learn the transformation parameters under the supervision of oriented bounding box (OBB) annotations, which is with lightweight and can be easily embedded into detectors for oriented object detection. In our work, our method is based on the RoI transformer to deal with OBB, we constrain random discarding and positive/negative confusion of small objects and produce qualified training samples with accurate locations and scopes without introducing complicated modules. Our experiments also confirm that it is unnecessary to preset a large number of specially designed anchors with large computational overheads.

\section{The Proposed Method}

In this section, we introduce the technical details of our proposed Dynamic Enhancement Anchor Network~(DEA-Net) and instantiate our Dynamic Enhancement Anchor~(DEA) module by showing how to apply the scheme to the object detectors with a feature pyramid for object detection in aerial images. Specifically, we first introduce the details of our proposed DEA module, and then introduce the sample discriminator which facilitates the learning of small objects. Then, we show the details of our overall network architecture. Finally, we show how to join training with inference of our DEA-Net.

\subsection{Dynamic Enhancement Anchor}\label{sec3.1}

\begin{figure}[t]
	\centering
	\includegraphics[width=1\linewidth]{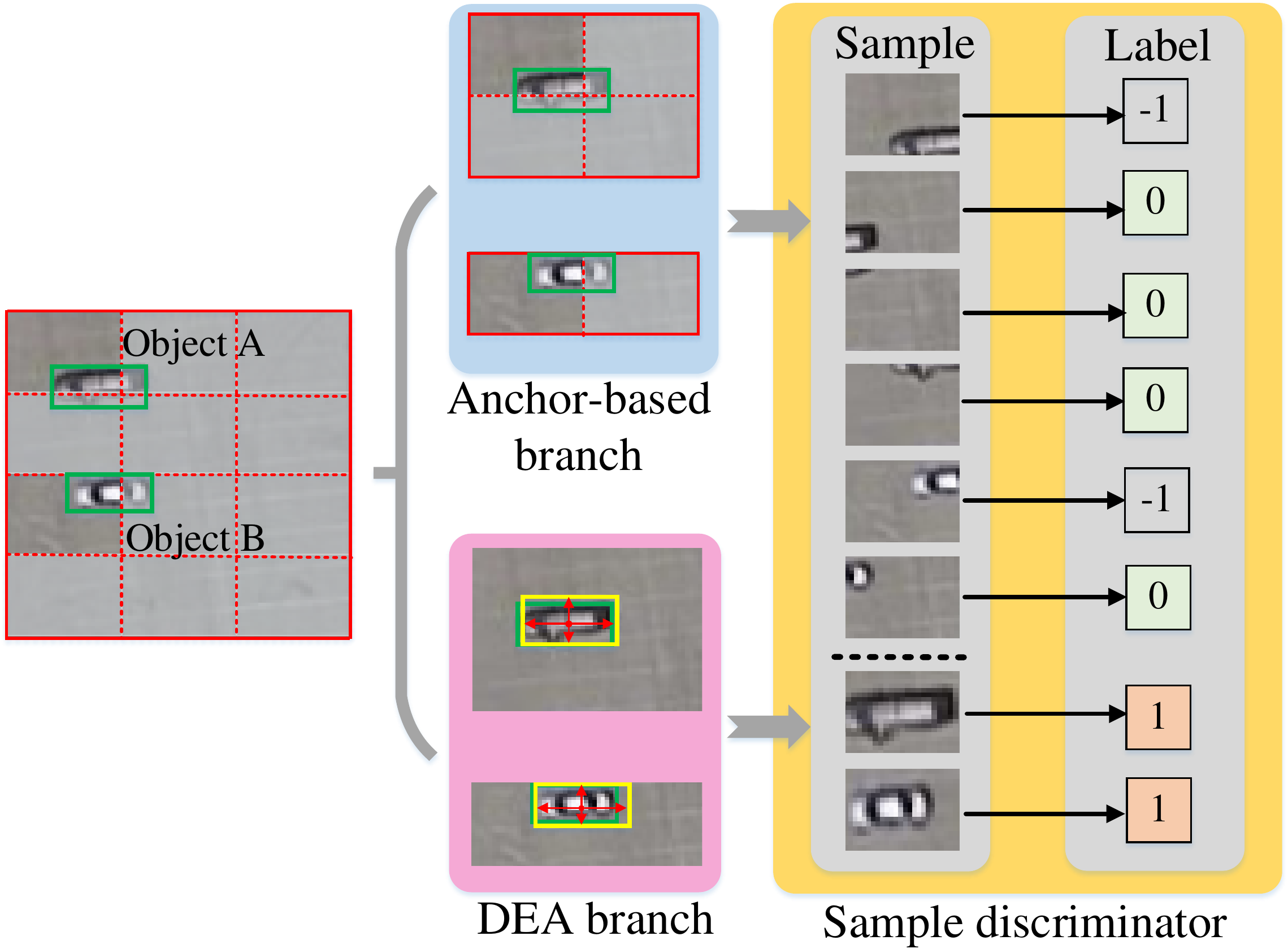}
	\caption{Dynamic enhancement anchor module to provide better positive samples of small objects. Specifically, in the original anchor-based module, the preset anchors are assigned as negative samples~($0$) or discarded samples ($-1$). However, our DEA module can generate high-quality bounding-boxes of higher Intersection-over-Union~(IoU), assigned as positive samples~($1$) for better training the network.}
	\label{method}
\end{figure}

In the literature, anchor-based methods find it difficult to fully learn small objects by selecting positive and negative samples for training through the examination of Intersection-over-Union~(IoU) overlap. As shown in Figure~\ref{method}, in the anchor-based module, if the IoU between the preset anchors and the ground-truth boxes of small objects is lower than the threshold of the positive samples, the samples are treated as discarded samples~(\ie, $-1$ as sample label) or negative samples~(\ie, $0$ as sample label). The training of anchor-based detectors for small objects is not sufficient because of the lack of positive samples, severely affecting the detection performance. 

As aforementioned, the regression bounding-boxes in the anchor-free detector usually have higher IoU for the small objects than the anchor-based detector. However, for some objects of a large size and extreme aspect ratios, anchor-free detectors may have poor performance. Therefore, instead of utilizing the anchor-free method to replace the anchor-based method to generate samples for training, we combine their advantages to deal with positive sample selection of different scales.

We construct an anchor-free sample generation module that shares the feature pyramid with the anchor-based module and integrate it with an anchor-based detector.  Via an interactive sample screening procedure in the sample discriminator, the anchor-free module is utilized to generate more positive samples of small objects which are ignored in the anchor-based module. For some objects of a large size and extreme aspect ratios, we preserve the anchors in the anchor-based module which have higher IoUs as the positive samples. 

\subsection{Interactive Sample Screening} 
Current studies~\cite{tian2019fcos, wang2018sface} have reported that the designed anchor boxes are the key to successful anchor-based detectors, and the detection performance is sensitive to the size, aspect ratio and the number of the anchor boxes. Therefore, the anchors in these anchor-based detectors must be carefully tuned for each specific task on different datasets. For example, to deal with the challenge of small object detection, one needs to design smaller anchors beforehand and densely locate them on the input image. This handling leads to extensive computational costs and the imbalanced problem. Therefore, our proposed dynamic enhancement method aims to use the preset number of the preset anchors to improve the detection performance of small objects with less computational cost. 

\begin{algorithm}[!t] 
\caption{Sample discriminator} 
\label{alg} 
\begin{algorithmic}[1] 
\REQUIRE ~~\\ 
$\mathcal{G}$ is the set of ground-truth boxes \\
$\mathcal{P}$ is the set of feature pyramid levels\\
$\mathcal{A}$ is the set of anchor boxes of the RPN outputs  \\
$\mathcal{V}$ is the set of predicted vector of anchor-free branch \\
$\mathcal{T_P}$ is the threshold of positive samples \\
$\mathcal{T_N}$ is the threshold of negative samples \\
\ENSURE ~~\\ 
$\mathcal{S_E}$ is the set of enhancement samples \\
$\mathcal{S_P}$ is the set of positive samples \\
$\mathcal{S_N}$ is the set of negative samples \\
\FOR{each ground-truth box $g \in \mathcal{G}$}
\FOR{each feature pyramid level $p_i \in \mathcal{P}$}
\STATE decoder predicted vector $\mathcal{V}$ to bounding-boxes $\mathcal{B}$: \\
$\mathcal{B} = Decoder(\mathcal{V})$;
\ENDFOR
\STATE calculate the Intersection-over-Union~(IoU) between $g$ and ${b_j} \in \mathcal{B}$: \\
$\mathcal{IB}_g = IoU(b_j,g)$;
\STATE calculate the Intersection-over-Union~(IoU) between $g$ and ${a_i} \in \mathcal{A}$: \\
$\mathcal{IA}_g = IoU(a_i,g)$;
\IF{$\mathcal{IB}_g^j \geq \mathcal{T_P}$ and $\mathcal{IB}_g^j \geq \mathcal{IA}_g^i$}
\STATE $\mathcal{S_E} = \mathcal{S_E} \bigcup \mathcal{B}_g^j$
\ELSIF{$\mathcal{IA}_g^i \geq \mathcal{T_P}$ and $\mathcal{IA}_g^i \geq \mathcal{IB}_g^j$}
\STATE $\mathcal{S_P} = \mathcal{S_P} \bigcup \mathcal{A}_g^i$
\ELSIF{$\mathcal{IA}_g^i \leq \mathcal{T_N}$}
\STATE $\mathcal{S_N} = \mathcal{S_N} \bigcup \mathcal{A}_g^i$
\ENDIF
\ENDFOR
\STATE $\mathcal{S_P} = \mathcal{S_P} \bigcup \mathcal{S_E}$
\RETURN $\mathcal{S_E}$, $\mathcal{S_P}$, $\mathcal{S_N}$;
\end{algorithmic}
\end{algorithm}
Algorithm~\ref{alg} describes how the proposed sample discriminator works with an input image. For each ground-truth box $g=[x,y,w,h,c]$ on the input image, where $(x,y)$ is the left-top corner of the box, $(w,h)$ are the box width and height, respectively, and $c$ is the class label, our anchor-free branch will generate prediction vectors: 
\begin{equation}
\mathcal{V} = [v_t^{m,n}, v_l^{m,n}, v_b^{m,n}, v_r^{m,n}, c^{m,n}], 
\end{equation}
where $v_t^{m,n}, v_l^{m,n}, v_b^{m,n}, v_r^{m,n}$ are the distances between the current pixel location $(m,n)$ and the top, left, bottom and right boundaries of the box, and $c^{m,n}$ is the prediction class label. We first decode the prediction vectors $\mathcal{V}$ to form the bounding-boxes $\mathcal{B}$:
\begin{equation}
\mathcal{B} = [x^{m,n},y^{m,n},w^{m,n},h^{m,n},c^{m,n}], 
\end{equation}
where $x^{m,n} = m - v_l^{m,n}$, $y^{m,n} = n - v_t^{m,n}$,\\ $w = v_l^{m,n} + v_r^{m,n}$, $h = v_t^{m,n} + v_b^{m,n}$. 

Then, we calculate the IoU between the regressed bounding-boxes $\mathcal{B}$ and the ground-truth $g$, labeled as $\mathcal{IB}_g$, and the IoU between the anchors of the RPN output in the anchor-based module $\mathcal{A}$ and the ground-truth $g$, labeled as $\mathcal{IA}_g$. Afterward, we select samples as follows: 

\begin{equation}
(\mathcal{IB}_g^j \geq \mathcal{IA}_g^i) \cap
(\mathcal{IB}_g^j \geq \mathcal{T_P}),
\end{equation}
where $\mathcal{T_P}$ is the threshold of the positive samples~(\ie, $0.5$ in this paper). It denotes the bounding-box in the anchor-free module have better consistency than the anchor proposal, and we assign box $\mathcal{B}_g^j$ to the enhanced samples $\mathcal{S_E}$ with
\begin{equation}
(\mathcal{IA}_g^i \geq \mathcal{T_P})\cap (\mathcal{IA}_g^i \geq \mathcal{IB}_g^j).
\end{equation}
We assign anchor $\mathcal{A}_g^i$ as positive samples $\mathcal{S_P}$. If we have
\begin{equation}
\mathcal{IA}_g^i \leq \mathcal{T_N},
\end{equation}
where $\mathcal{T_N}$ is the threshold of the negative samples~(\ie, $0.3$ in this paper), we assign anchors $\mathcal{A}_g^i$ as negative samples $\mathcal{S_N}$. Finally, we add the enhancement samples $\mathcal{S_E}$ to the positive samples $\mathcal{S_P}$.

\begin{figure}[t]
	\centering
	\includegraphics[width=1\linewidth]{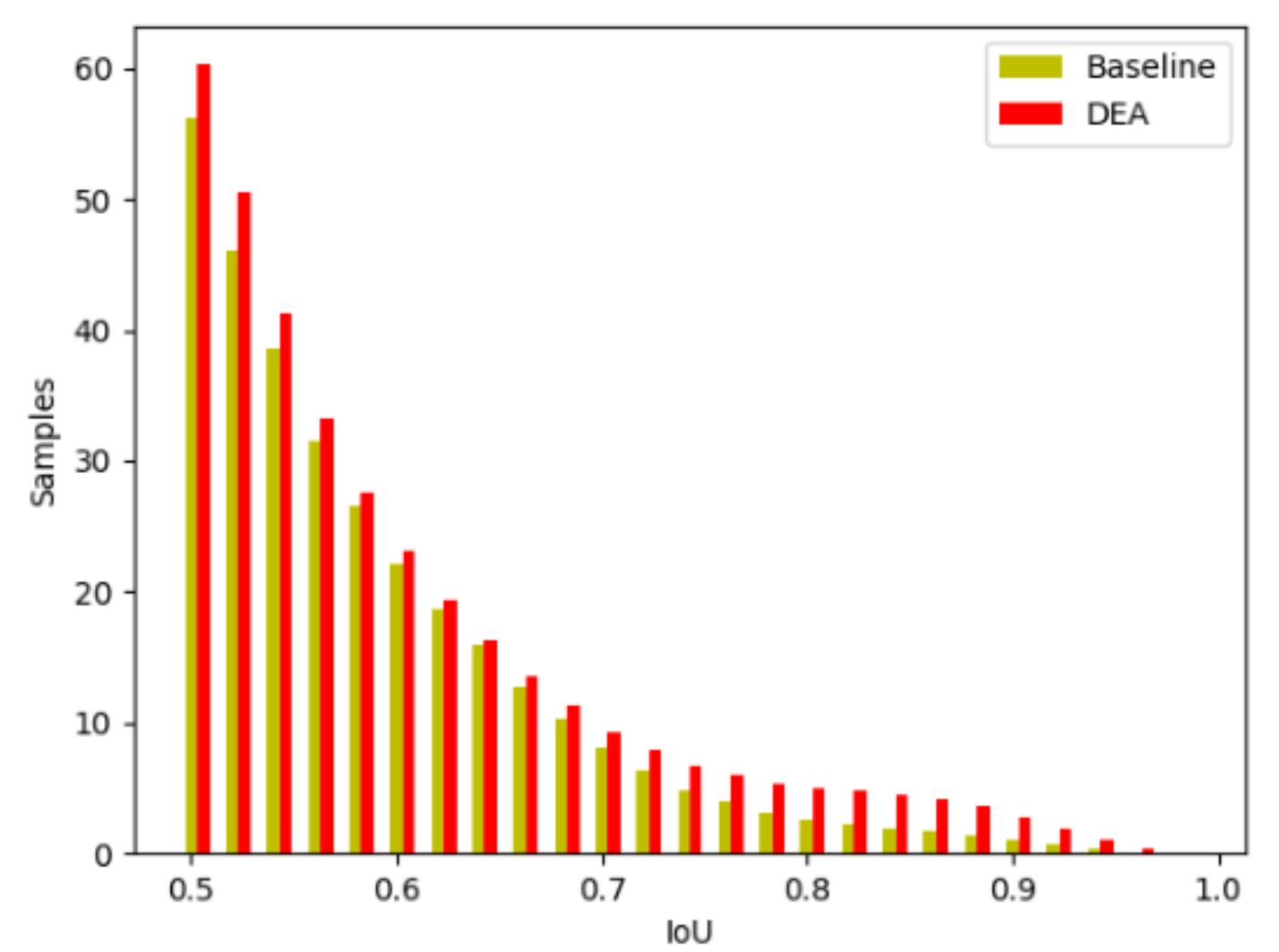}
	\caption{The Intersection-over-Union~(IoU) distributions of the baseline~(Faster R-CNN~\cite{ren2016faster}) and our DEA-net. The $x$-axis represents the IoU between the samples and the ground-truth boxes. The $y$-axis represents the average number of samples on each image.}
	\label{statistic}
\end{figure}
The threshold of positive and negative samples affects the accuracy of the detection. Because if the threshold is higher, the numbers of positive samples will decrease. If the threshold is lower, the quality of the samples will decrease. We keep this hyperparameter that is widely used in the baseline just like Faster R-CNN, and we find that doing so would potentially cause the problem of insufficient positive samples. The introduction of DEA alleviates the above-mentioned risks, and it directly and independently generates more qualified positive samples.  As shown in Figure~\ref{statistic}, we study the IoU distributions of the samples generated by Faster R-CNN~\cite{ren2016faster} and our DEA-net. Statistics shows that our DEA network provides more positive samples with higher IoUs compared with the original anchor-based detector. 

In our method, the positive samples dynamically generated by the DEA module are all horizontal bounding box, which are the same as the anchor preset in the anchor-based branch. Then, we leverage a sample discriminator to realize interactive sample screening between the anchor-based branch and DEA branch to generate eligible samples. Finally, we utilize RoI-Transformer~\cite{ding2019learning} to obtain the feature of rotated objects.

\begin{table*}[!htb]
	\begin{center}
	\renewcommand\tabcolsep{2.6pt}
	\begin{threeparttable}
		\scalebox{1.05}{
			\begin{tabular}{cccc|ccccccccccccccc|c}
				\hline
				\hline
				 R-50 & R-101 & R-152 & +DEA & PL & BD & BR & GTF & SV & LV & SH & TC & BC & ST & SBF & RA & HA & SP & HC & mAP (\%)\\
				\hline
				\checkmark & & & & 88.25 & 77.05 & 51.94 & 64.98 & 77.98 & 76.83 & 87.02 & 90.78 & 82.75 & 82.84 & 55.56 & 62.70 & 73.92 & 67.74 & 58.59  & 73.26\\
				
				 \checkmark & & & \checkmark& 87.97	& 79.14 & 51.13 & 65.43 & 79.87 & 78.24 & 87.66 & 90.59 & 83.28 & 85.65 & 53.72 & 63.39 & 73.83 & 69.50 & 57.32 & \textbf{73.78}$_{\color{red}{ +0.52 }}$ \\
					\hline
				 & \checkmark  & & & 88.53 & 77.70 & 51.59 & 68.80 & 74.02 & 76.85 & 86.98 & 90.24 & 84.89 & 77.68 & 53.91 & 63.56 & 75.88 & 69.48 & 55.50 &   73.06\\
				
				  & \checkmark & & \checkmark&88.32 & 79.18 & 52.03 & 69.50 & 78.21 & 77.98 & 87.76 &90.21  & 85.12 & 83.53 & 54.35 & 62.08 & 73.52 & 70.62 & 56.94  &  \textbf{73.96}$_{\color{red}{ +0.90 }}$ \\
					\hline
				 & &\checkmark & & 88.56 & 77.71 & 54.03 & 72.76 & 74.15 & 77.48 & 87.17 & 90.17 & 76.39 & 83.95 & 45.68 & 64.50 & 76.22 & 69.53 & 53.41  & 72.78\\
				
				 &  & \checkmark& \checkmark& 88.43	& 79.21 & 51.28 & 69.46 & 78.17 & 79.19 & 87.21 & 89.89 & 78.20 & 85.98 & 45.94 & 63.56 & 74.77 & 70.97 & 55.83 & \textbf{73.21}$_{\color{red}{ +0.43 }}$ \\
				\hline
				\hline
			\end{tabular}
		}
		\end{threeparttable}
	\end{center}
	\caption{The effectiveness of our proposed method with different backbone networks on the test set of DOTA~\cite{xia2018dota} for oriented object detection. ``+ DEA'' indicates the implementation of our proposed module on the backbone networks.}
	\label{t1}
\end{table*}

\subsection{Network Architecture}\label{sec3.2}

We build a Dynamic Enhancement Anchor network~(DEA-Net) for both oriented and horizontal object detection tasks in aerial images. Figure~\ref{structure} illustrates the overall architecture of our DEA module integrated with Faster RCNN~\cite{ren2016faster} and RoI-Transformer~\cite{ding2019learning} for oriented object detection. We deploy ResNet~\cite{he2016deep} as the backbone, which has been pre-trained on the ImageNet~\cite{deng2009imagenet}. Then, we construct a multi-scale feature pyramid~\cite{lin2017feature} in the top-down pathway from the backbone network with levels from $\mathcal{P}_2$ to $\mathcal{P}_6$ and $\mathcal{P}_i$ has $1/2^i$ resolution of the input image. Then, we construct a DEA head to $\mathcal{P}_i$, which contains the anchor-based module and our proposed DEA module. We construct the anchor-based module following the technique reported in ~\cite{ren2016faster}, including the RPN head network to generate horizontal region proposals.

For the DEA module, we construct an anchor-free module following the approach shown in~\cite{tian2019fcos}. We add four convolutional layers after the feature maps $\mathcal{P}_i$ created by the standard FPN for classification, centralization and regression.  We decode the prediction vectors of the anchor-free module to form bounding-boxes and then we select better positive samples from the two modules through the sample discriminator. Finally, for the task of oriented object detection, we build the rotated head inspired by RoI-Transformer~\cite{ding2019learning} which transforms the horizontal proposals to the rotated ones for arbitrary-oriented detection and a standard Faster R-CNN~\cite{ren2016faster} is used for horizontal object detection. 
The anchor-free and anchor-based modules work jointly in a multi-task style and share the features at each pyramid level.

A recent work for improving one-stage detectors is to introduce an individual prediction branch to estimate the quality of localization, where the predicted quality facilitates the classification to improve detection performance~\cite{li2020generalized}. The authors compared IoU-branch and centerness-branch, and believed that IoU-branch performs consistently better than centerness as a measurement of localization quality. The convincing reason is that centerness scores are much smaller than IoU scores, which cause the final scores of bounding boxes are potentially small and then removed by NMS. In our method, we utilize DEA-branch (an anchor-free branch with centerness loss) to assist the training process of the anchor-based detector to generate eligible training sample according to the ground truth sample rather than generating the final output score. This avoids the divergence of the two branches in the inference stage. We introduce centerness loss as a part of loss for training the DEA branch, and we select samples based on the IoU between regressed bounding-boxes and the ground-truth, not based on the centerness score.

\subsection{Training and Inference}\label{sec3.3}
\myparagraph{Multi-task joint training} 

Integrated with Faster RCNN~\cite{ren2016faster}, our DEA module is trained jointly with the anchor-based module in a multi-task style, as shown in Figure~\ref{structure}. We define $\mathcal{L}_{ab}$ as the total loss of the anchor-based module, and $\mathcal{L}_{af}$ as the total loss of the anchor-free module. We combine the losses from the anchor-based and anchor-free modules as the loss of the entire network. Then, the total optimization loss for the whole network is 
\begin{equation}
\mathcal{L} = \mathcal{L}_{ab} + \mathcal{L}_{af},
\end{equation}

For the multi-task loss in the anchor-based detection module, following~\cite{ren2016faster}, we optimize the target of the detection by regressing anchor boxes. The loss function for each anchor can be formulated as:
\begin{equation}
\mathcal{L}_{ab}(\{p_i\},\{t_i\}) = \mathcal{L}_{ab\_cls}(p_i, p_i^*) + {p_i^*} \mathcal{L}_{ab\_reg}(t_i, t_i^*)
\end{equation}
where the classification loss $\mathcal{L}_{ab\_cls}$ is the cross entropy loss, $p_i$ is the predicted probability of anchor $i$ being an object and $p_i^*$ represents its ground-truth label~($p_i^* = 1$ for positive samples and $p_i^* = 0$ for negative samples). The regression loss $\mathcal{L}_{ab\_reg}$ is smooth L1 loss~\cite{girshick2015fast}, $t_i$ is the vector of the predicted box and $t_i^*$ represents the ground-truth box.

For the anchor-free module, following~\cite{tian2019fcos}, the loss function for each location can be formulated as:
\begin{equation}
\begin{aligned}
\mathcal{L}_{af}(\{p_{m,n}\},\{t_{m,n}\})=\mathcal{L}_{af\_cls}(p_{m,n}, p_{m,n}^*) \\
+\mathds{1}_{\left\{p_{m, n}^{*}>0\right\}} \mathcal{L}_{af\_reg}(t_{m,n}, t_{m,n}^*)\\
+\mathds{1}_{\left\{p_{m, n}^{*}>0\right\}}\mathcal{L}_{af\_center}(t_{m,n}, {t}_{m, n}^{*})
\end{aligned}
\end{equation}
where classification loss $\mathcal{L}_{af\_cls}$ is focal loss~\cite{lin2017focal}, $p_{m,n}$ is the prediction of class labels and $p_{m,n}^*$ represents the ground-truth label. The regression loss $\mathcal{L}_{af\_reg}$ is IoU loss~\cite{yu2016unitbox}. $\mathds{1}_{\left\{p_{m, n}^{*}>0\right\}}$ is the indicator function, being $1$ if $p_{m, n}^{*}>0$ and $0$ otherwise. $t_{m,n}$ is a vector of the predicted box and $t_{m,n}^*$ represents the ground-truth. The center-ness loss $\mathcal{L}_{af\_center}$ is the cross entropy loss.

\begin{table}[!htb]
	\begin{center}
	\renewcommand\tabcolsep{2.6pt}
	\begin{threeparttable}
		\scalebox{1.2}{
			\begin{tabular}{cccc|cc}
				\hline
				\hline
				 R-50 & R-101 & R-152 & +DEA  & GFLOPs / FPS & $\#$Params (M) \\
				\hline
				\checkmark & & & &  211.30 / 14.8 & 55.13 6\\
				
				 \checkmark & & & \checkmark&  225.21 / 12.5 &  59.90\\
					\hline
				 & \checkmark  & & &   289.19 / 12.7 & 74.12\\
				
				  & \checkmark & & \checkmark& 303.10 / 10.6  & 78.89 \\
					\hline
				 & &\checkmark & &   367.17 / 11.1 &  89.77 \\
				
				 &  & \checkmark& \checkmark&  381.08 / 9.2 & 94.54\\
				\hline
				\hline
			\end{tabular}
		}
		\end{threeparttable}
	\end{center}
	\caption{The efficiency of our proposed method with different backbone networks on the test set of DOTA~\cite{xia2018dota} for oriented object detection. ``+ DEA'' indicates the implementation of our proposed module on the backbone networks.}
	\label{t11}
\end{table}

\begin{table*}[!htb]
	\begin{center}
	\renewcommand\tabcolsep{2.6pt}
	\begin{threeparttable}
		\scalebox{0.9}{
			\begin{tabular}{ccc|ccccccccccccccc|cc|c}
				\hline
				\hline
				 Baseline & +Anchor & +DEA & PL & BD & BR & GTF & SV & LV & SH & TC & BC & ST & SBF & RA & HA & SP & HC & GFLOPs / FPS & $\#$Params (M) & mAP (\%) \\
				\hline

				 \checkmark &  &  & 88.53 & 77.70 & 51.59 & 68.80 & 74.02 & 76.85 & 86.98 & 90.24 & 84.89 & 77.68 & 53.91 & 63.56 & 75.88 & 69.48 & 55.50 & 289.19 / 12.7 & 74.12 &73.06\\
				 
				 \checkmark & \checkmark & & 88.41 & 80.14 & 53.79 & 70.70 & 77.82 & 76.98 & 86.98 & 90.75 & 83.90 & 81.13 & 51.95 & 61.41 & 74.89 & 69.15 & 59.27 & 310.86 / 9.5 & 74.13 & 73.82\\
				
				  \checkmark&  & \checkmark & 88.32 & 79.18 & 52.03 & 69.50 & 78.21 & 77.98 & 87.76 &90.21  & 85.12 & 83.53 & 54.35 & 62.08 & 73.52 & 70.62 & 56.94 & 303.10 / 10.6 & 78.89 & \textbf{73.96}\\
				
				\hline
				\hline
			\end{tabular}
		}
		\end{threeparttable}
	\end{center}
	\caption{The comparison of efficiency between our proposed method and the method of preset more small anchors on DOTA~\cite{xia2018dota} for oriented object detection. ``Baseline'' indicates the Faster RCNN with the backbone of ResNet-101. ``+ Anchor'' indicates the implementation of more small anchors on the baseline networks. ``+ DEA'' indicates the implementation of our proposed module on the baseline networks.}
	\label{t111}
\end{table*}

\myparagraph{Inference with anchor-based module} 

Our DEA-Net utilizes the anchor-free and the anchor-based modules to jointly train the network to strengthen its feature representation ability and provide high-quality samples to the training task. During the inference stage, we feed the images to the anchor-based module whilst freezing the anchor-free module. This is mainly due to the fact that the anchor-free module has relatively poor consistency in locating bounding-boxes, especially for the objects of a large aspect ratio. Freezing the anchor-free module could also avoid complicated fusion computation to control the computational overhead for the inference. We use the confidence score $0.05$ and set the threshold of non-maximum suppression to be $0.1$ to generate the final detection results. We demonstrate the effectiveness of the proposed scheme in the following ablation experiments.

\myparagraph{Discussion} 

In essence, our method is still anchor-based two-stage detection pipeline. 
One critical problem we solve is the “anchor over-quantization” problem, which would cause small targets to be ignored in the anchor laying process, thereby breaking the original statistical distribution of training samples and finally affecting the performance of the trained model.
Instead of increasing the anchor laying density (which would greatly increase training overhead), we design a sample discriminator in the training stage. Unlike the solutions that combine anchor-based and anchor-free methods to detect + fusion, the proposed sample discriminator (as shown in Figure~\ref{structure} and Algorithm~\ref{alg}) comprehensively evaluates the consistency of anchor-boxes (produced in the anchor-based branch) and the inferred box produced by the DEA branch (an anchor-free branch) with ground truth. As demonstrated in Figure~\ref{method}, small targets that are split into negative samples by anchor boxes are completely retained in the DEA module, free from the quantization errors of anchors. In the sample discriminator, these target regions located by the DEA module are further regarded as positive samples to compensate for quantization errors in the anchor-based branch, as shown in Figure~\ref{statistic}.

\section{Experimental work}
\subsection{Settings}
\myparagraph{Datasets} 

\emph{{DOTA}}~\cite{xia2018dota} is one of the large datasets for object detection in aerial images with both oriented and horizontal bounding box annotations. It contains $2,806$ aerial images with $188,282$ annotated instances from different sensors and platforms. The image size ranges from around $800 \times 800$ to $4,000 \times 4,000$ pixels and contains objects exhibiting in a wide variety of scales, orientations, and shapes. DOTA contains $15$ object categories, including Plane (PL), Baseball diamond (BD), Bridge (BR), Ground track field (GTF), Small vehicle (SV), Large vehicle (LV), ship (SH), Tennis court (TC), Basketball court (BC), Storage tank (ST), Soccer-ball field (SBF), Roundabout (RA), Harbor (HA), Swimming pool (SP), and Helicopter (HC). In our experiments, following~\cite{xia2018dota,yang2019scrdet}, $3/6$ of the original images are randomly selected to form the training set, $1/6$ as the validation set, and $2/6$ as the testing set.
\emph{{HRSC2016}}~\cite{lb2017high} is a challenging dataset for ship detection in aerial images with large aspect ratios and arbitrary orientations. These images were collected from Google Earth, which contain $1061$ images and more than $20$ categories of ships in various appearances. The image size ranges from $300 \times 300$ to $1500 \times 900$. In our work, following~\cite{lb2017high}, the training, validation, and test sets include $436$, $181$, and $444$ images, respectively. For HRSC2016, only oriented object detection can be carried out. 

\myparagraph{Image size} 

For DOTA and HRSC2016, we generate a series of $1,024 \times 1,024$ patches from the original images with a stride of $824$ for training, validation, and testing.

\myparagraph{Baseline setup}

We use the standard two-stage detector Faster R-CNN~\cite{ren2016faster} as the baseline. It utilizes ResNet-101 as backbone. FPN~\cite{lin2017feature} is adopted to construct a feature pyramid. Predefined horizontal anchors are set on each feature level, \ie, $P$2 - $P$6. Here, we do not use any rotation anchor. For oriented object detection, we add the rotated head developed in RoI-Transformer~\cite{ding2019learning} which transforms the horizontal proposals to the rotated ones. For a fair comparison, all the experimental data and parameter settings are strictly consistent as those reported in~\cite{ding2019learning, xia2018dota, lb2017high}.

To verify the universality of our approach, we also embed our approach to ReDet~\cite{han2021redet} which incorporates rotation-equivariant networks into the detector to extract rotation-equivariant features. It uses the ReResNet-50~\cite{han2021redet} as backbone, and FPN~\cite{lin2017feature} is adopted to construct a feature pyramid. And then it also adds the rotated head developed in RoI-Transformer~\cite{ding2019learning} for arbitrary-oriented detection.

\myparagraph{Hyper-parameters}

For the hyper-parameters, following~\cite{ding2019learning, ming2020dynamic}, in DOTA and HRSC2016, only three horizontal anchors are set with the aspect ratios of \{$1/2$, $1$, $2$\}, the base anchor scale is set as \{$8^2$\}, and the anchor strides of each level of the feature pyramid are set to be \{$4$, $8$, $16$, $32$, $64$\}. 

For the positive and negative sample selection, following~\cite{ren2016faster, ding2019learning}, we set the threshold of the positive samples as $\mathcal{T_P} = 0.5$ and the threshold of the negative samples as $\mathcal{T_N} = 0.3$.

We set $\gamma = 2$ and $\alpha = 0.25$ for the focal loss in $\mathcal{L}_{af\_cls}$. 

\myparagraph{Implementation details}

In order to verify the effectiveness of our method, we perform ablation studies on the DOTA dataset, and avoid utilizing any bells-and-whistles training strategy and data augmentation in the ablation study. 

For the peer comparison on DOTA and HRSC2016, like~\cite{ding2019learning, yang2019scrdet, ming2020dynamic}, we only conduct rotation augmentation using $4$ angles~($0, 90, 180,270$) to simply avoid the imbalance between different categories.

Stochastic gradient descent is used as the optimizer. The initial learning rate is set to $0.005$ and divided by $10$ at each decay step. Weight decay and momentum are set to $0.0001$ and $0.9$, respectively. Following~\cite{xia2018dota, lb2017high}, the total iterations for DOTA and HRSC2016 are 80k and 20k, respectively. We train the models on RTX 2080Ti with a batch size of $1$.

\myparagraph{Evaluation and metrics}

Following~\cite{xia2018dota}, the standard mean Average Precision (mAP) is used as the primary evaluation metric for accuracy. Moreover, to verify the model efficiency, the model Parameters ($\#$Params), and GFLOPs / FPS are also taken into consideration. The results of DOTA reported in our work are obtained by submitting our predictions to the official DOTA evaluation server\footnote{https://captain-whu.github.io/DOTA/}.

\begin{table*}[t]
    \begin{center}
    \renewcommand\tabcolsep{4.0pt}
    \begin{threeparttable}
        \scalebox{1.2}{
            \begin{tabular}{c|c|c}
                \hline
                \hline
               Inference schemes & FPS & mAP (\%) \\
                \hline
                 Anchor-based baseline~Faster RCNN~\cite{ren2016faster} & 13.0 & 73.89\\
                Anchor-free baseline~FCOS~\cite{tian2019fcos}& 11.0 & 68.45\\
                \hline
                \textbf{Ours (DEA-Net) freezing anchor-free branch} & 10.8 & 74.85$_{\color{red}{ +0.96/+6.40 }}$\\
                Ours (DEA-Net) anchor-free + anchor-based fusion& 6.2 & 74.96$_{\color{red}{ +1.07/+6.51}}$\\
                \hline
                \hline
            \end{tabular}
        }
    \end{threeparttable}
    \end{center}
    \caption{The effectiveness of different inference schemes with our DEA-Net on  DOTA~\cite{xia2018dota} for horizontal object detection in aerial images. ResNet-101~\cite{he2016deep} is the backbone.}
    \label{t2}
\end{table*}

\subsection{Ablation Study}

Our ablation study is carried out on DOTA~\cite{xia2018dota} for oriented object detection with ResNet-101~\cite{he2016deep}, which aims to: (1) verify the effectiveness of our method on different backbone networks; (2) verify the effectiveness of our proposed units integrated with the baseline; (3) verify the effectiveness of the inference schemes.

\begin{table*}[!htb]
	\begin{center}
	\renewcommand\tabcolsep{2.5pt}
	\begin{threeparttable}
		\scalebox{.97}{
			\begin{tabular}{r|c|ccccccccccccccc|c}
				\hline
				\hline
				Methods & Backbone & PL & BD & BR & GTF & SV & LV & SH & TC & BC & ST & SBF & RA & HA & SP & HC & mAP (\%) \\
				\hline
				\multicolumn{18}{c}{{\textbf{Oriented object detection}}}\\
				\hline
				FR-O~\cite{xia2018dota} (CVPR 2018) & R-101 & 79.09 & 69.12 & 17.17 & 63.49 & 34.20 & 37.16 & 36.20 & 89.19 & 69.60 & 58.96 & 49.40 & 52.52 & 46.69 & 44.80 & 46.30 & 52.93 \\
				
				R-DFPN~\cite{yang2018automatic} (ISO4 2018)& R-101 & 80.92 & 65.82 & 33.77 & 58.94 & 55.77 & 50.94 & 54.78 & 90.33 & 66.34 & 68.66 & 48.73 & 51.76 & 55.10 & 51.32 & 35.88 & 57.94 \\
				
				R$^2$CNN~\cite{jiang2017r2cnn} (preprint 2017) & R-101 & 80.94 & 65.67 & 35.34 & 67.44 & 59.92 & 50.91 & 55.81 & 90.67 & 66.92 & 72.39 & 55.06 & 52.23 & 55.14 & 53.35 & 48.22 & 60.67 \\
				
				RRPN~\cite{ma2018arbitrary} (TMM 2018)& R-101 & 88.52 & 71.20 & 31.66 & 59.30 & 51.85 & 56.19 & 57.25 & 90.81 & 72.84 & 67.38 & 56.69 & 52.84 & 53.08 & 51.94 & 53.58 & 61.01 \\
				
				ICN~\cite{azimi2018towards} (ACCV 2018)& R-101 & 81.36 & 74.30 & 47.70 & 70.32 & 64.89 & 67.82 & 69.98 & 90.76 & 79.06 & 78.02 & 53.64 & 62.90 & 67.02 & 64.17 & 50.23 & 68.16 \\
				
    			RoI Trans~\cite{ding2019learning} (CVPR 2019)& R-101 & 88.64	& 78.52 & 43.44 & \textbf{75.92} & 68.81 & 73.68 & 83.59 & 90.74 & 77.27 & 81.46 & 58.39 & 53.54 & 62.83 & 58.93 & 47.67 & 69.56 \\
				
				CAD-Net~\cite{zhang2019cad} (TGRS 2019) & R-101 & 87.80 & 82.40 & 49.40 & 73.50 & 71.10 & 63.50 & 76.70 & \textbf{90.90} & 79.20 & 73.30 & 48.40 & 60.90 & 62.00 & 67.00 & 62.20 & 69.90 \\
				
				DRN~\cite{pan2020dynamic} (CVPR 2020)& H-104 & 88.91	& 80.22 & 43.52 & 63.35 & 73.48 & 70.69 & 84.94 & 90.14 & 83.85 & 84.11 & 50.12 & 58.41 & 67.62 & 68.60 & 52.50 & 70.70 \\
				
				O$^2$-DNet~\cite{wei2020oriented} (ISPRS 2020)& H-104 & 89.31	& 82.14 & 47.33 & 61.21 & 71.32 & 74.03 & 78.62 & 90.76 & 82.23 & 81.36 & 60.93 & 60.17 & 58.21 & 66.98 & 61.03 & 71.04 \\
				
				SCRDet~\cite{yang2019scrdet} (ICCV 2019) & R-101 & 89.98	& 80.65 & 52.09 & 68.36 & 68.36 & 60.32 & 72.41 & 90.85 & \textbf{87.94} & 86.86 & 65.02 & 66.68 & 66.25 & 68.24 & 65.21 & 72.61 \\
				
				R$^3$Det~\cite{yang2019r3det} (preprint 2019)& R-152 & 89.49	& 81.17 & 50.53 & 66.10 & 70.92 & 78.66 & 78.21 & 90.81 & 85.26 & 84.23 & 61.81 & 63.77 & 68.16 & 69.83 & 67.17 & 73.74 \\
				
				CSL~\cite{yang2020arbitrary} (ECCV 2020) & R-152 & \textbf{90.25}	& \textbf{85.53} & 54.64 & 75.31 & 70.44 & 73.51 & 77.62 & 90.84 & 86.15 & 86.69 & \textbf{69.60} & \textbf{68.04} & 73.83 & 71.10 & 68.93 & 76.17 \\
				
				DAL~\cite{ming2020dynamic} (AAAI 2021)& R-50 & 89.69	& 83.11 & 55.03 & 71.00 & 78.30 & 81.90 & \textbf{88.46} & 90.89 & 84.97 & \textbf{87.46} & 64.41 & 65.65 & \textbf{76.86} & 72.09 & 64.35 & 76.95 \\
				
				R$^3$Det-DCL~\cite{yang2020dense} (CVPR 2021)& R-152 & 89.26 & 83.60 & 53.54 & 72.76 & 79.04 & 82.56 & 87.31 & 90.67 & 86.59 & 86.98 & 67.49 & 66.88 & 73.29 & 70.56 & \textbf{69.99} & 77.37\\

				\textbf{Ours (RoI Trans + DEA)} & R-101 & 89.18	& 83.46 & \textbf{55.14} & 71.69 & \textbf{79.59} & \textbf{83.08} & 88.10 & 90.88 & 87.09 & 86.73 &63.99 & 65.14 &75.81 & \textbf{78.01} & 68.69 & \textbf{77.77}$_{\color{red}{+0.40}}$\\
				
				\hline
				\multicolumn{18}{c}{\textbf{Horizontal object detection}}\\
				\hline
				SSD~\cite{liu2016ssd} (ECCV 2016)& R-101 & 44.74 	& 11.21 & 6.22 & 6.91 & 2.00 & 10.24 & 11.34 & 15.59 & 12.56 & 17.94 & 14.73 & 4.55 & 4.55 & 0.53 & 1.01 & 10.94 \\
				
				YOLOv2~\cite{redmon2016you} (CVPR 2016) & R-101 & 76.90 & 33.87 & 22.73 & 34.88 & 38.73 & 32.02 & 52.37 & 61.65 & 48.54 & 33.91 & 29.27 & 36.83 & 36.44 & 38.26 & 11.61 & 39.20 \\
				
				R-FCN~\cite{dai2016r} (NIPS 2016)& R-101 & 79.33	& 44.26 & 36.58 & 53.53 & 39.38 & 34.15 & 47.29 & 45.66 & 47.74 & 65.84 & 37.92 & 44.23 & 47.23 & 50.64 & 34.90 & 47.24 \\
				
				FR-H~\cite{xia2018dota} (CVPR 2018)& R-101 & 80.32 & 77.55 & 32.86 & 68.13 & 53.66 & 52.49 & 50.04 & 90.41 & 75.05 & 59.59 & 57.00 & 49.81 & 61.69 & 56.46 & 41.85 & 60.46 \\
				
				FPN~\cite{lin2017feature} (CVPR 2017)& R-101 & 88.70	& 75.10 & 52.60 & 59.20 & 69.40 & 78.80 & 84.50 & 90.60 & 81.30 & 82.60 & 52.50 & 62.10 & \textbf{76.60} & 66.30 & 60.10 & 72.00 \\
				
				ICN~\cite{azimi2018towards} (ACCV 2018) & R-101 & 90.00	& 77.70 & 53.40 & \textbf{73.30 }& 73.50 & 65.00 & 78.20 & 90.80 & 79.10 & 84.80 & 57.20 & 62.10 & 73.50 & 70.20 & 58.10 & 72.50 \\
				
				SCRDet~\cite{yang2019scrdet} (ICCV 2019) & R-101 &\textbf{ 90.18} & 81.88 & 55.30 & 73.29 & 72.09 & 77.65 & 78.06 & \textbf{90.91} & 82.44 & 86.39 & \textbf{64.53} & 63.45 & 75.77 & 78.21 & 60.11 & 75.35 \\
				
				\textbf{Ours (RoI Trans + DEA)} & R-101 & 89.18 & \textbf{83.34} & \textbf{58.94} & 71.69 &\textbf{80.23} & \textbf{83.97} & \textbf{88.26} & 90.88 & \textbf{87.09} & \textbf{87.44} & 64.24 & \textbf{65.04} & 76.40 & \textbf{80.88} & \textbf{68.81} & \textbf{78.43}$_{\color{red}{+3.08}}$\\
				\hline
				\hline
			\end{tabular}
		}
		\end{threeparttable}
	\end{center}
	\caption{Comparisons with other state-of-the-art methods on the test set of DOTA~\cite{xia2018dota} for both oriented and horizontal object detection in aerial images. ``Ours'' means the implementation of the DEA module on the baseline model. ``R-'' in the Backbone column denotes ResNet~\cite{he2016deep}, and ``H-'' denotes the Hourglass network~\cite{newell2016stacked}.}
	\label{t3}
\end{table*}

\begin{table*}[!htb]
	\begin{center}
	\renewcommand\tabcolsep{2.5pt}
	\begin{threeparttable}
		\scalebox{.97}{
			\begin{tabular}{r|c|ccccccccccccccc|c}
				\hline
				\hline
				Methods & Backbone & PL & BD & BR & GTF & SV & LV & SH & TC & BC & ST & SBF & RA & HA & SP & HC & mAP (\%) \\
				\hline
				S$^2$A-Net~\cite{han2021align} (TGRS 2021) & R-50 & 88.89 & 83.60 & 57.74 & 81.95 & 79.94 & 83.19 & \textbf{89.11} & 90.78 & 84.87 & 87.81 & 70.30 & \textbf{68.25} & 78.30 & 77.01 & 69.58 & 79.42 \\
				
				ReDet~\cite{han2021redet} (CVPR 2021) & ReR-50 & 88.81 & 82.48 & \textbf{60.83} & 80.82 & 78.34 & 86.06 & 88.31 & \textbf{90.87} & 88.77 & 87.03 & 68.65 & 66.90 & \textbf{79.26} & \textbf{79.71} & 74.67 & 80.10\\
				
				GWD~\cite{yang2021rethinking} (ICML 2021) & R-152 & 89.66 & \textbf{84.99} & 59.26 & \textbf{82.19} & 78.97 & 84.83 & 87.70 & 90.21 & 86.54 & 86.85 & \textbf{73.47} & 67.77 & 76.92 & 79.22 & 74.92 & 80.23\\
				
				\textbf{Ours (ReDet + DEA)} & ReR-50 &  \textbf{89.92} & 83.84 & 59.65 & 79.88 & \textbf{80.11} & \textbf{87.96} & 88.17 & 90.31 & \textbf{88.93} & \textbf{88.46} & 68.93 & 65.94 & 78.04 & 79.69 & \textbf{75.78} & \textbf{80.37}$_{\color{red}{+0.14}}$ \\ 
				\hline
				\hline
			\end{tabular}
		}
		\end{threeparttable}
	\end{center}
	\caption{Comparisons with some current state-of-the-art methods on the test set of DOTA~\cite{xia2018dota} for oriented object detection in aerial images. ``Ours'' means the implementation of the DEA module on the baseline model. ``R-'' in the Backbone column denotes ResNet~\cite{he2016deep}, and ``ReR-'' denotes Rotation-equivariant ResNet~\cite{han2021redet}.}
	\label{t33}
\end{table*}

\begin{figure*}[!htb]
\centering
\includegraphics[width=.95\textwidth]{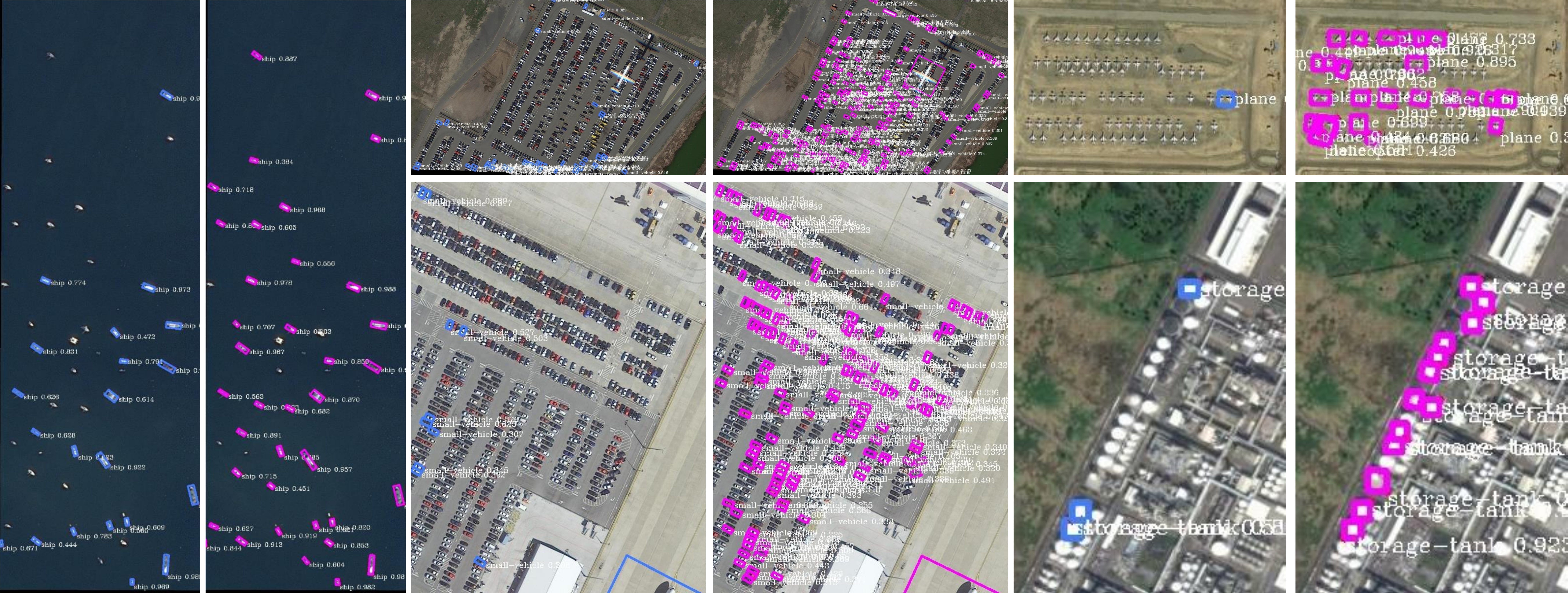}
\caption{Comparison against the baseline (Faster RCNN~\cite{ren2016faster}+RoI-Transformer~\cite{ding2019learning}) on DOTA~\cite{xia2018dota} for oriented object detection with ResNet-101~\cite{he2016deep}. ${\color{blue}{\textrm{Blue~boxes}}}$ indicate the results of the baseline and ${\color{magenta}{\textrm{pink~boxes}}}$ are the results of our proposed DEA-Net.}
\label{Figure_show_compare}
\end{figure*}

\begin{figure*}[!htb]
	\begin{center}
	\includegraphics[width=.95\linewidth]{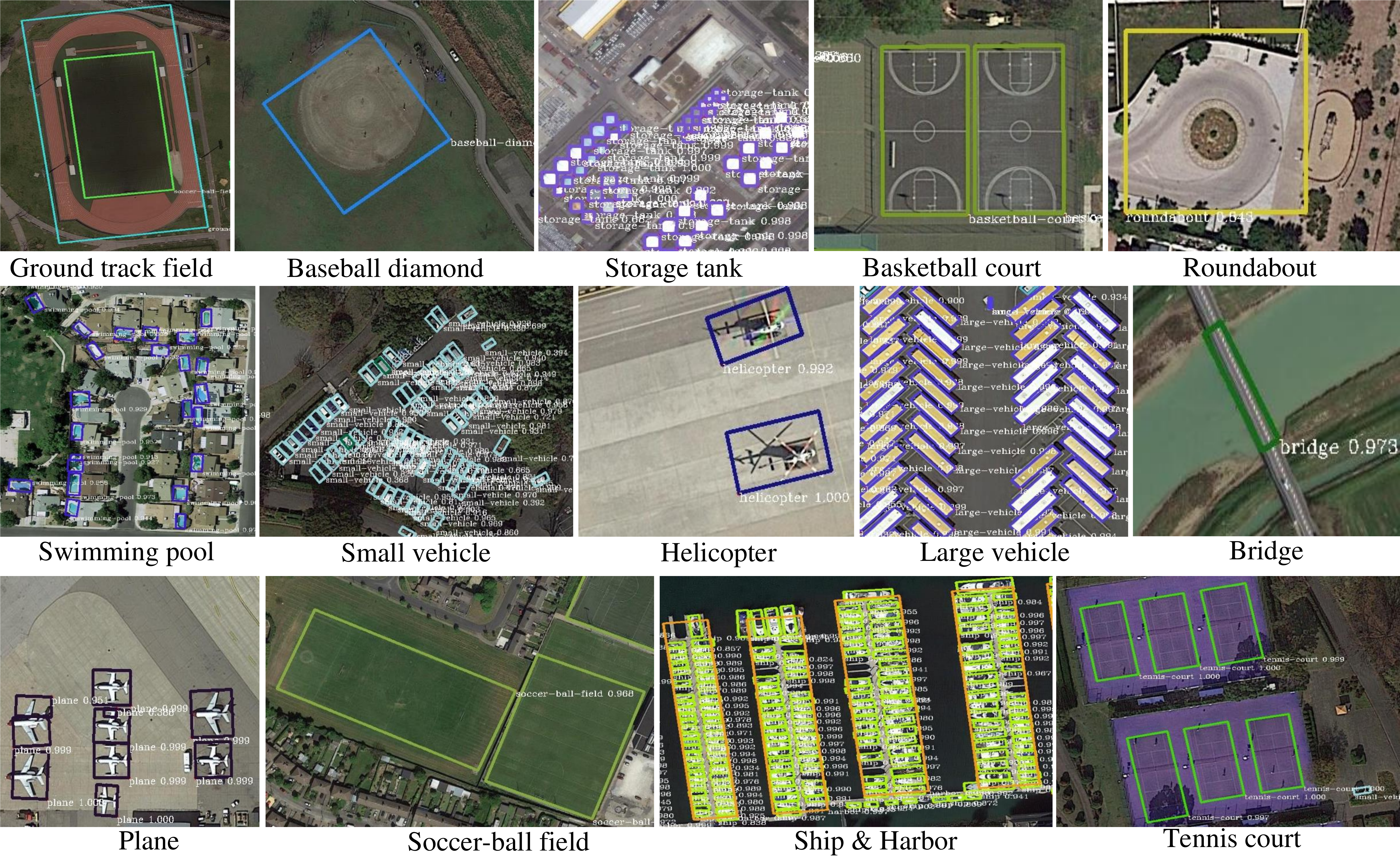}
	\caption{Visualization results for oriented object detection on the test set of DOTA~\cite{xia2018dota}}
	\label{show_all}
	\end{center}
\end{figure*}
\myparagraph{Effectiveness and efficiency on different backbones} 

In Table~\ref{t1}, we show the experimental results of different backbone networks with our proposed units on the test set of DOTA. We use mean Average Precision~(mAP) to examine our proposed module with ResNet-50, ResNet-101, and ResNet-152, respectively. Note that for aerial image, the object detection using oriented bounding box (OBB) is much more important but more difficult than using horizontal bounding box (HBB), that is why in Table ~\ref{t1} we perform ablation study on OBB task rather than HBB. We observe that adding our proposed module to the backbone increases mAP by $0.52\%$, $0.90\%$, and $0.43\%$. 

In Table~\ref{t11}, we report the model $\#$Params and GFLOPs / FPS for the evaluation of model efficiency. It is clear that using our proposed DEA module increases a little computational cost. For example, average increases on these three backbones are around $4.47$ M model $\#$Params with around $13.91$ GFLOPs, and with around $2$ FPS reduction. Considering the model performance and the amount of the calculation, in the following experiments, we select ResNet-101 as our backbone network. 

\myparagraph{Effectiveness of the proposed units}

In Table~\ref{t1}, we show the performance of our DEA module integrated with three backbones for $15$ categories of DOTA. We witness that our module can bring improvements for the bounding box mAP by $0.90\%$. We can observe that our module has large improvements on small objects. Specifically, for SV~(Small vehicle), our method can increase AP by $4.19\%$, and for ST~(Storage Tank), AP can be increased by $5.85\%$. 

\myparagraph{Efficiency of our proposed units}

We also compare the efficiency and effectiveness of our proposed method against those of the method of presetting more small anchors, as shown in Table~\ref{t111}. The Hyper-parameters settings of the comparison experiment are as follows: \newline
(1) The base anchor scale of the baseline and our method is \{$8^2$\}, and we set the base anchor scale of the method of presetting more small anchors as \{$2^2$,$4^2$,$8^2$\}. \newline
(2) The aspect ratios of our method and the method of presetting more small anchors are both \{$1/2$, $1$, $2$\}. \newline
(3) The anchor strides of each feature map of the two methods are both \{$4$, $8$, $16$, $32$, $64$\}. 

It is clear that the method of presetting more anchors increases more computational cost than the proposed method. The GFLOPs increase of $21.67$ (+Anchor) \emph{vs} $13.91$ (+DEA), with the FPS reduction of $3.2$ (+Anchor) \emph{vs} $2.1$ (+DEA). This is because in the method of presetting more anchors, the number of anchors has tripled and these anchors would participate in the calculation of the horizontal bounding boxes and the rotated bounding boxes. In terms of the performance of the two methods, the method of presetting more anchors would indeed improve the performance of some small objects~(\eg, small vehicle). Our method has more improvements on small objects, because our DEA module can dynamically generate positive samples which match objects better.

\myparagraph{Effectiveness of the inference schemes}

We also compare the effectiveness of different inference schemes on DOTA for horizontal object detection after having trained the anchor-based and anchor-free baselines with our proposed method, shown in Table~\ref{t2}. Compared with the two baselines, our DEA-Net of freezing the anchor-free module can increase mAP by $0.96\%$ and $6.40\%$. When we fuse the outputs of the anchor-based and anchor-free modules, mAP can have a minor improvement ($0.96\%$ \emph{vs} $1.07\%$ and $6.40\%$ \emph{vs} $6.51\%$), compared with the former. Meanwhile, fusing the two modules for inference, the inference speed becomes clearly slower ($10.8$ FPS \emph{vs} $6.2$ FPS). That is why after having trained the DEA-net, we freeze the anchor-free branch and only utilize the anchor-based module for inference.      

\myparagraph{Visualizations}

\begin{figure}[!htb]
	\centering
	\includegraphics[width=1\linewidth]{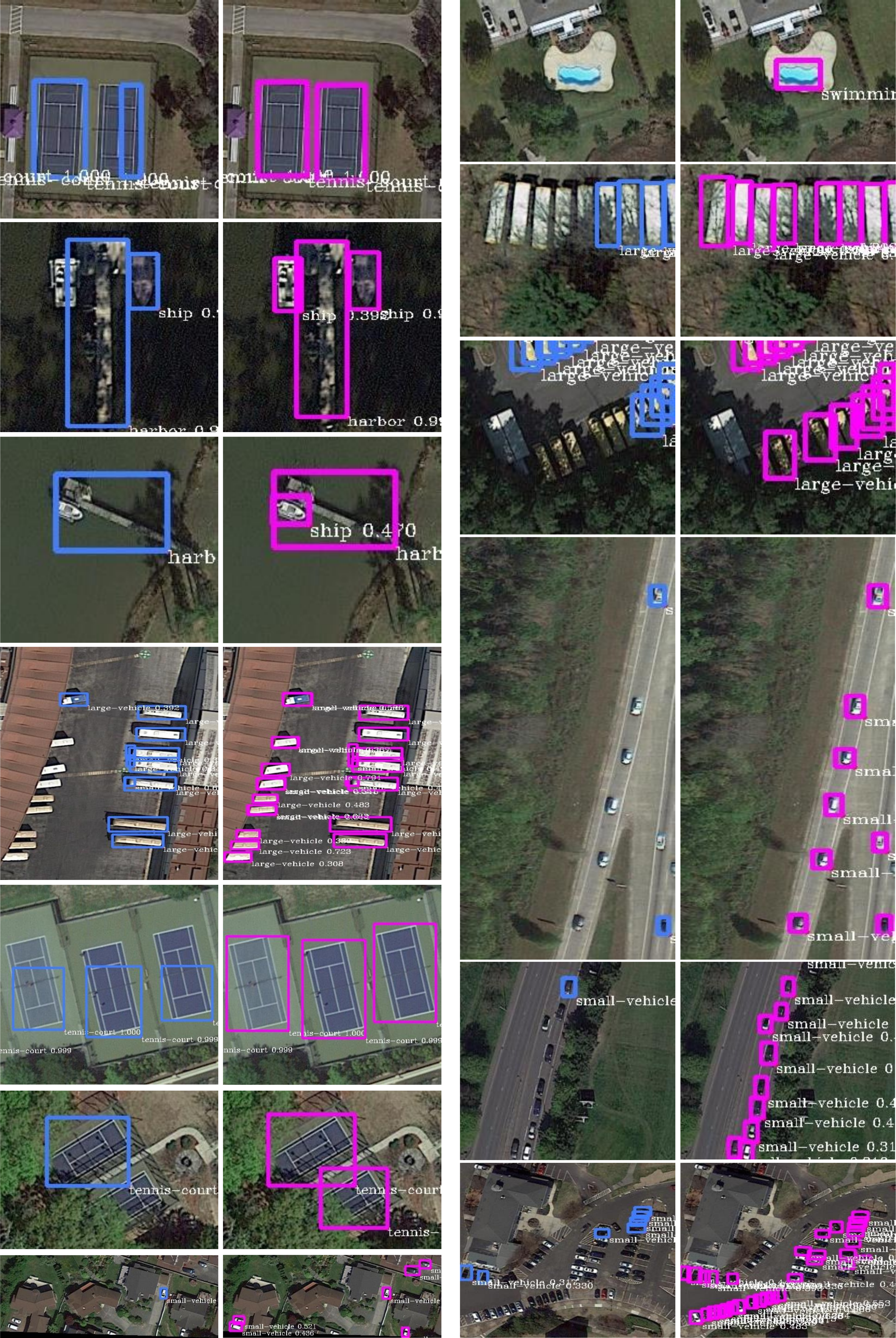}
	\caption{Some visualized comparisons of object detection with horizontal bounding box on DOTA. The figures with ${\color{blue}{\textrm{blue~boxes}}}$ are the results of the baseline (Faster R-CNN) and ${\color{magenta}{\textrm{pink~boxes}}}$ are the results of our proposed DEA-Net.}
	\label{view_DOTA}
\end{figure}

We show some of the visual comparisons for oriented object detection between the baseline and the proposed method in Figure~\ref{Figure_show_compare}. The proposed method achieves notably better precision for small object detection, such as small vehicles, storage tank, ship, and airplane. We also show some visualized comparisons for horizontal object detection on DOTA in Figure~\ref{view_DOTA}. Our proposed method also achieves better performance both in the case of small and large objects with horizontal bounding boxes.

\subsection{Comparisons with state-of-the-arts}\label{sec4.3}
\begin{table}[t]
	\begin{center}
	\renewcommand\tabcolsep{5.0pt}
	\begin{threeparttable}
		\scalebox{1.1}{
			\begin{tabular}{r|c|ccccccccccccccc|c}
				\hline
				\hline
				Methods & Backbone & mAP (\%) \\
				\hline
				R$^2$CNN~\cite{jiang2017r2cnn} (preprint 2017)& R-101 & 73.07\\
				RCI\&RC2~\cite{lb2017high} (ICPRAM 2017) & V-16 & 75.70\\
				RRPN~\cite{ma2018arbitrary} (TMM 2018)& R-101 & 79.08\\
				R$^2$PN~\cite{zhang2018toward} (GRSL 2018)& V-16 & 79.60\\
				RRD~\cite{liao2018rotation} (CVPR 2018)& V-16 & 84.30\\
				RoI Trans~\cite{ding2019learning} (CVPR 2019)& R-101 & 86.20\\
				Gliding Vertex~\cite{xu2020gliding} (TPAMI 2020) & R-101 & 88.20\\
				R-RetinaNet~\cite{lin2017focal} (ICCV 2017)& R-101 & 89.18\\
				R$^3$Det~\cite{yang2019r3det} (preprint 2019)& R-101 & 89.26\\
				RetinaNet-DAL~\cite{ming2020dynamic} (AAAI 2021)& R-101 & 89.77\\
				R$^3$Det-DCL~\cite{yang2020dense} (CVPR 2021)& R-101 & 89.46\\
				\textbf{Ours (RoI Trans + DEA)} & R-101 & \textbf{90.56}$_{\color{red}{ +1.10 }}$\\
				\hline
				\hline
			\end{tabular}
		}
	\end{threeparttable}
	\end{center}
	\caption{Comparisons with other state-of-the-art methods on the test set of HRSC2016~\cite{lb2017high} for oriented object detection in aerial images. ``R-'' in the Backbone column denotes the ResNet~\cite{he2016deep}, and ``V-'' denotes the VGG network~\cite{simonyan2014very}. mAP is obtained on the VOC 2007 evaluation metric.}
	\label{t4}
\end{table}

\myparagraph{Results on DOTA}

We compare the proposed approach with some state-of-the-art methods on the test set of DOTA, as shown in Table~\ref{t3}. When our approach integrated with RoI-Transformer~\cite{ding2019learning}, our DEA-Net achieves $77.77\%$ mAP for oriented object detection and $78.43\%$ mAP for horizontal object detection, and outperforms many advanced methods. Of these $15$ categories, DEA-Net ranks at the top for $4$ categories for oriented object detection and $10$ categories for horizontal object detection. Moreover, DEA-Net surpasses the advanced method by 0.40 mAP for oriented object detection with a weaker backbone network (ResNet-101 vs ResNet-152) and 3.08 mAP for horizontal object detection with the same backbone. Visualization results on the test set of DOTA are shown in Figure~\ref{show_all}. DEA-Net can accurately predict the categories and have satisfactory performance on small objects, such as small vehicle, storage tank and ship.

We also compare the proposed approach with some newest methods on the test set of DOTA, as shown in Table~\ref{t33}. To verify the universality of our approach, we embed our approach to one of these current detectors ReDet~\cite{han2021redet}, which is a state-of-the-art rotation detector that explicitly encodes rotation equivariance and rotation invariance. We integrate our DEA module with ReDet and conduct data augmentation following the way in \cite{han2021redet} (\ie, multi-scale data and random rotation), our method achieves  $80.37\%$ mAP for oriented object detection, and of these $15$ categories, it ranks at the top for $5$ categories.

\myparagraph{Results on HRSC2016} 

\begin{figure}[!htb]
	\centering
	\includegraphics[width=1\linewidth]{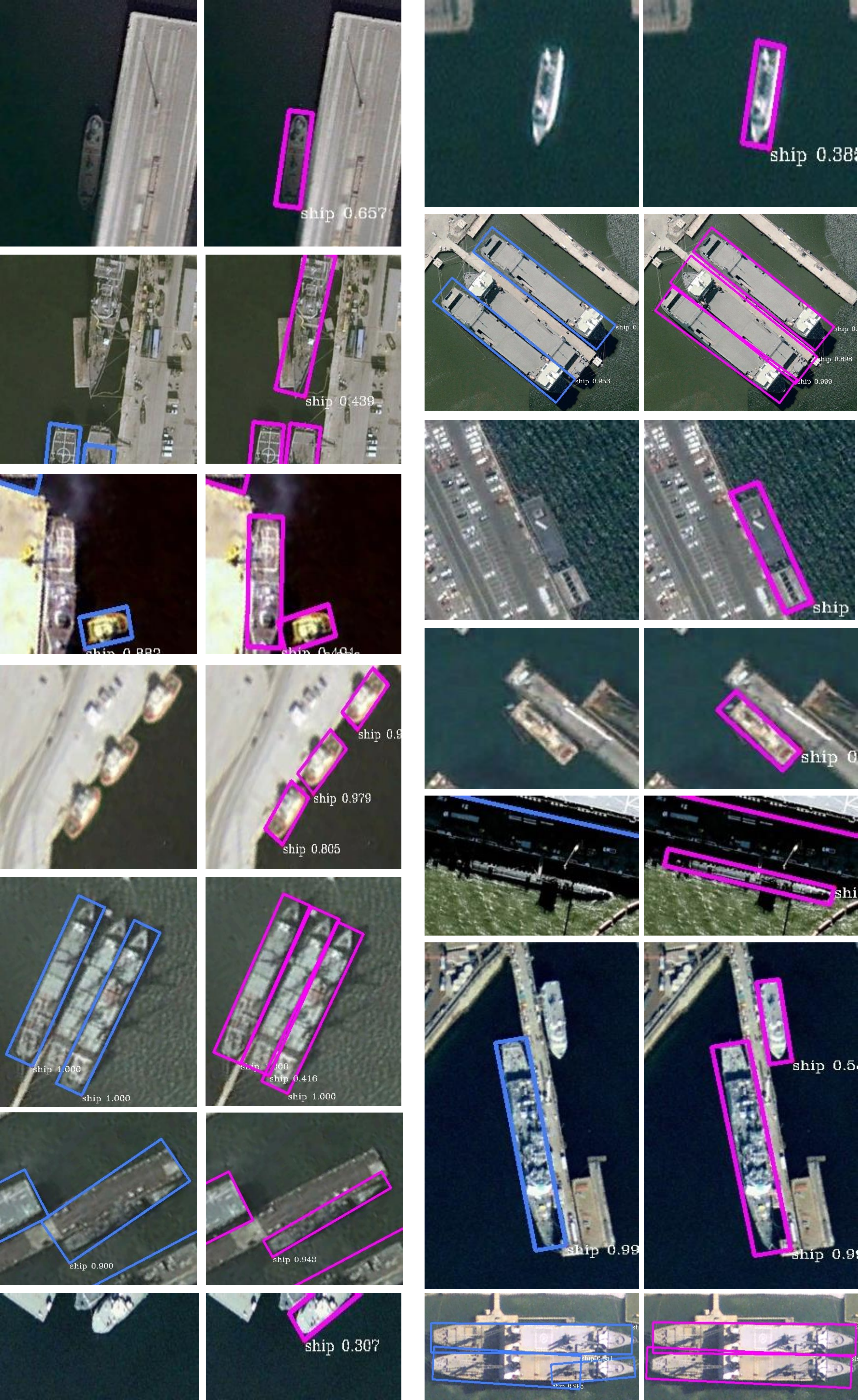}
	\caption{Some visualized comparison of object detection with oriented bounding box on HRSC2016. The figures with ${\color{blue}{\textrm{blue~boxes}}}$ are the results of the baseline (RoI-Transformer + Faster R-CNN) and ${\color{magenta}{\textrm{pink~boxes}}}$ are the results of our proposed DEA-Net.} 
	\label{view_HRSC}
\end{figure}

The comparisons with the other state-of-the-art methods on the test set of HRSC2016~\cite{lb2017high} are shown in Table~\ref{t4}. We can observe that our method achieves the state-of-the-art performance in mAP by $90.56\%$, which surpasses the previous best model by $1.1\%$. Particularly, in our experiments, our DEA-Net uses only $3$ horizontal anchors with the aspect ratios of \{$1/2$, $1$, $2$\}, but outperforms the other frameworks with a large number of anchors. We show some of the visual comparisons for oriented object detection between the baseline and the proposed method on HRSC2016 in Figure~\ref{view_HRSC}. Our proposed method also achieves better precision for objects with large aspect ratio. The experiments show that it is critical to effectively utilize the predefined anchors and select high-quality samples where our DEA module can regress the bounding boxes at the locations of the objects without presetting a large number of rotated anchors. 

\myparagraph{Discussion} 

The object detection using oriented bounding box (OBB) is much more difficult than using horizontal bounding box (HBB). Table~\ref{t3} and Table~\ref{t4} also support this phenomenon. In Table~\ref{t3}, for OBB task the proposed methods outperform SOTA by 0.4\% but for HBB it outperforms SOTA by 3.08\%. In Table~\ref{t3} and Table~\ref{t4}, we can also find that, for OBB task, recent SOTA method can have only $<$ 1\% gain compared with the previous state of the art methods, for example, R3Det-DCL ~\cite{yang2020dense} (CVPR 2021) 77.37\% vs DAL ~\cite{ming2020dynamic} (AAAI 2021) 76.95\% on DOTA.

\begin{figure}[!htb]
	\centering
	\includegraphics[width=1\linewidth]{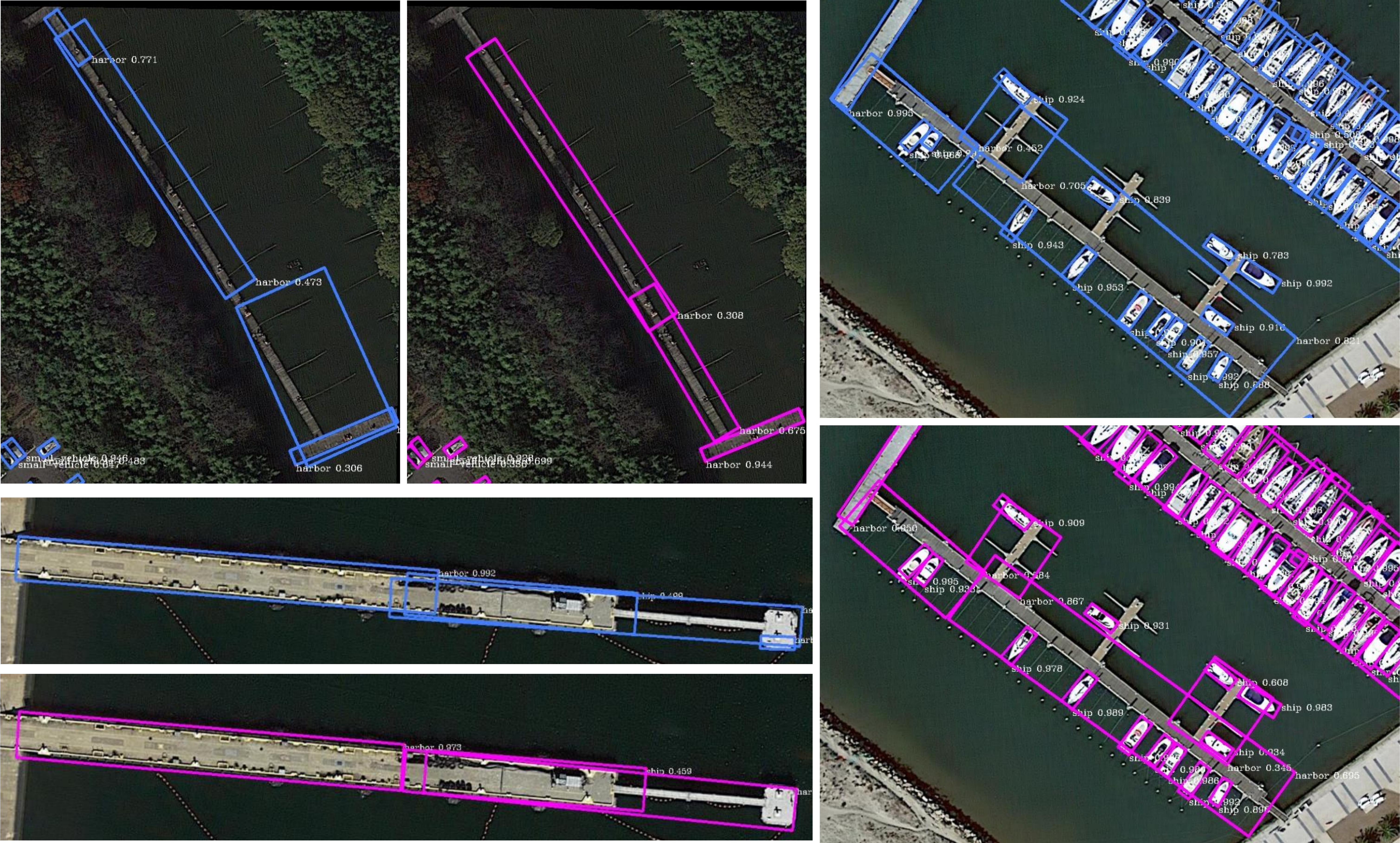}
	\caption{Some failed visualized comparisons of baseline and our method. The figures with ${\color{blue}{\textrm{blue~boxes}}}$ are the results of the baseline (Faster R-CNN + RoI-Transformer) and ${\color{magenta}{\textrm{pink~boxes}}}$ are the results of the proposed DEA-Net.}
	\label{view_failed_case}
\end{figure}

We also list some failed visualization comparisons of baseline and our method, as shown in Figure~\ref{view_failed_case}. The results show that our method, similar to the baseline, cannot generate accurate bounding boxes when detecting objects with extreme aspect ratios~(\eg, harbor). This may be because the DEA module fail to generate accurate positive samples when the objects are of a large size and an extreme aspect ratio, while the anchor-based branch~(\eg, the baseline) also cannot regress accurate bounding boxes when the aspect ratio of preset anchors are not match the objects of extreme aspect ratio.
 
\section{Conclusions and future work}
In this work, a simple yet effective DEA module was proposed to facilitate the learning of small objects. We implemented our DEA module on the standard object detection backbone network with a feature pyramid network (\ie, DEA-Net) and conducted extensive experiments on both oriented and horizontal object detection in aerial images. Experimental results on the challenging DOTA and HRSC2016 indicated that our proposed DEA-Net could achieve state-of-the-art performance in accuracy with moderate computational overhead.

In the future, we will extend the proposed DEA-Net to a broader range of natural scenes. Besides, exploring how to use DEA-Net for semantic and panoramic segmentation is also a promising direction.

\bibliographystyle{IEEEtran}
\bibliography{dea}
\clearpage

\appendices
\section{Some visualized comparison of object detection with oriented bounding box on HRSC2016.}


\newpage
\section{Some visualized comparisons of object detection with horizontal bounding box on DOTA. }

\section*{Acknowledgment}

The authors would like to thank Professor Sheng-Jun Huang from NUAA and Professor Gui-Song Xia from Wuhan University for their help discussion and comments. This work was supported by AI+ Project of NUAA (XZA20003), Natural Science Foundation of China (62172218, 61772268, 62172212), Natural Science Foundation of Jiangsu Province (BK20190065). 

\ifCLASSOPTIONcaptionsoff
  \newpage
\fi

\end{document}